%% file: main.tex
\documentclass{article}

\usepackage{microtype}
\usepackage{graphicx}
\usepackage{subfigure}
\usepackage{booktabs} 
\usepackage[algo2e, ruled]{algorithm2e}

\usepackage{hyperref}

\usepackage[accepted]{icml2024}

\usepackage{amsmath}
\usepackage{amssymb}
\usepackage{mathtools}
\usepackage{amsthm}
\usepackage[capitalize,noabbrev]{cleveref}

\usepackage{multirow}
\usepackage{makecell}
\theoremstyle{plain}

\theoremstyle{definition}

\theoremstyle{remark}

\usepackage[textsize=tiny]{todonotes}

\usepackage{tcolorbox}

\newtcolorbox[auto counter,number within=section]{myquote}{colframe=blue,colback=blue!10,boxrule=0.5pt,boxsep=5pt,left=5mm,right=5mm,top=2mm,bottom=2mm}

\newcommand{\algo}{{\textit{DynSyn}}}
\newcommand{\model}{{\small \sc\textsf{MS-Human-700}}}

\icmltitlerunning{Dynamical Synergistic Representation for Efficient Learning and Motor Control}

\begin{document}

\twocolumn[
\icmltitle{\algo: Dynamical Synergistic Representation for Efficient Learning \\ and Control in Overactuated Embodied Systems}




\begin{icmlauthorlist}
\icmlauthor{Kaibo He}{comp}
\icmlauthor{Chenhui Zuo}{comp}
\icmlauthor{Chengtian Ma}{comp}
\icmlauthor{Yanan Sui}{comp}
\end{icmlauthorlist}

\icmlaffiliation{comp}{School of Aerospace Engineering, Tsinghua University, Beijing, China}
\icmlcorrespondingauthor{Yanan Sui}{ysui@tsinghua.edu.cn}

\icmlkeywords{Reinforcement Learning, Motor Control, Overactuated Systems, Synergistic Representation}

\vskip 0.3in
]



\printAffiliationsAndNotice{}  

\begin{abstract}
Learning an effective policy to control high-dimensional, overactuated systems is a significant challenge for deep reinforcement learning algorithms. Such control scenarios are often observed in the neural control of vertebrate musculoskeletal systems. The study of these control mechanisms will provide insights into the control of high-dimensional, overactuated systems. The coordination of actuators, known as muscle synergies in neuromechanics, is considered a presumptive mechanism that simplifies the generation of motor commands. The dynamical structure of a system is the basis of its function, allowing us to derive a synergistic representation of actuators. Motivated by this theory, we propose the \textit{Dynamical Synergistic Representation} (\algo) algorithm. \algo\ aims to generate synergistic representations from dynamical structures and perform task-specific, state-dependent adaptation to the representations to improve motor control. We demonstrate \algo's efficiency across various tasks involving different musculoskeletal models, achieving state-of-the-art sample efficiency and robustness compared to baseline algorithms. \algo \ generates interpretable synergistic representations that capture the essential features of dynamical structures and demonstrates generalizability across diverse motor tasks. 
\end{abstract}

\input{sections/introduction}

\input{sections/relatedwork}

\input{sections/dynamical_feature}

\input{sections/dynamical_synergies}

\input{sections/experiments}
\input{sections/results}
\input{sections/conclusion}
\input{sections/impact}

\bibliography{References}
\bibliographystyle{icml2024}

\newpage
\appendix
\onecolumn

\input{sections/appendix}

\end{document}

%% file: sections/introduction.tex
\section{Introduction}

In the evolution of embodied intelligence, researchers have used reinforcement learning (RL) algorithms to train controllers across diverse robotic platforms, yielding notable advancements in motor control. These agents can acquire robust and generalizable policies through iterative trial and error within large-scale simulations, subsequently deploying them onto real-world robots via sim-to-real methodologies \cite{rudin2022learning, duan2022survey, radosavovic2023learning}. Overactuation and redundancy can often enhance the safety and robustness of embodied intelligent systems, mitigating the risk of sudden control failures \cite{hsu1989dynamic, schneiders2004benefits, tohidi2016fault}. However, overactuation will increase the complexity of the controlled object, particularly by enlarging the dimensionality of the action space, making it difficult for the deep RL controllers to achieve motor control.

\begin{figure*}[htbp]
  \begin{center}
    \includegraphics[width=1\linewidth]{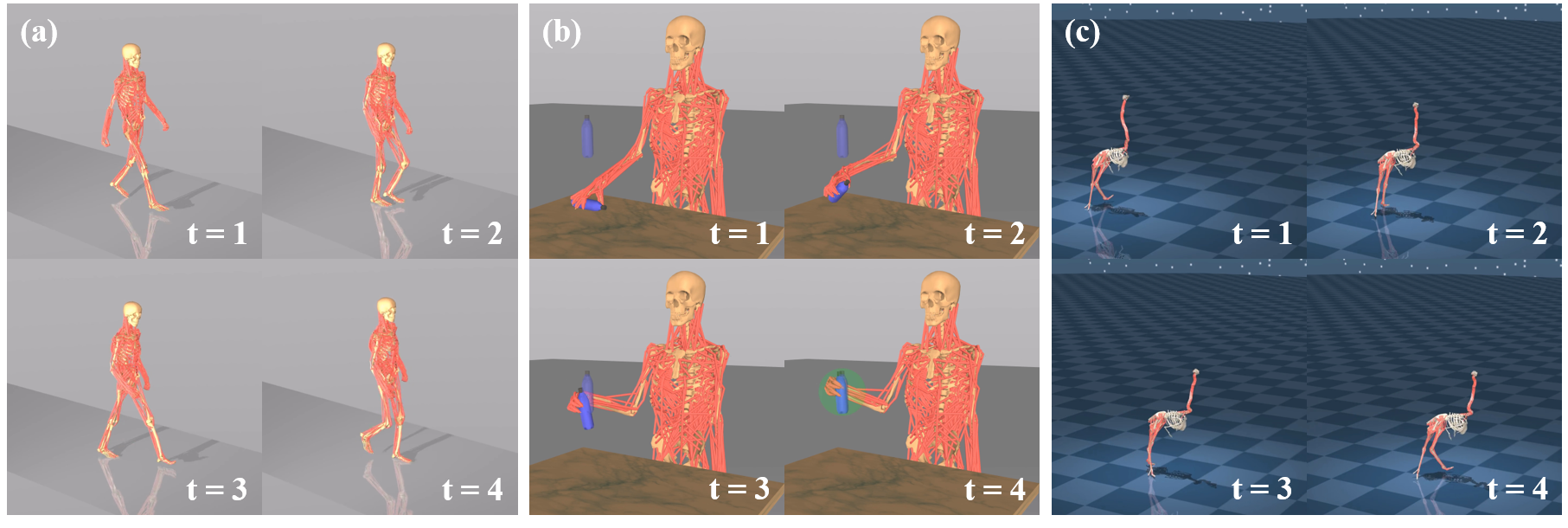}
  \vskip -0.1in
  \caption{\textbf{Motor behaviors of overactuated musculoskeletal systems acquired by \algo.} (a) Gait of \model~model. (b) Manipulation of \model~Arm model. (c) Locomotion of \textit{Ostrich} model. See our anonymous project website at \url{https://sites.google.com/view/dynsyn}. }
 
  \label{fig:motor_behaviors}
  \end{center}
  \vskip -0.1in
\end{figure*}

A common example of overactuated embodied systems is musculoskeletal systems in nature. In contrast to existing DRL agents, the motor control intelligence of vertebrates can control musculoskeletal systems through the central nervous system, exhibiting the ability to generalize across a variety of motor tasks while maintaining stability even under large disturbances, such as external force interference and drastic changes of environmental parameters. Exploring musculoskeletal motion control can help address the control challenges posed by high-dimensional, overactuated systems, thereby advancing our progress towards embodied intelligence. Nevertheless, a musculoskeletal model of human possesses characteristics that pose significant challenges for motor control by a RL agent. Firstly, the muscle control parameter space is high-dimensional, where over 600 skeletal muscles control hundreds of joints \cite{bernstein1966co}. Secondly, the system is overactuated, as multiple muscles actuate one joint and multiple joints may be affected by one muscle \cite{ting2012review}. Thirdly, the dynamic of neuro-muscular actuators is non-linear and inconstant \cite{zajac1989muscle, wolpert2000computational}, and these actuators can only generate tension and no reverse forces. 

How does motor control in vertebrates effectively address the challenge of redundant actuation? In neuroscience, there is a hypothesis known as muscle synergies. It proposes that coordinated recruitment of groups of muscles serves as a modular framework for biological motor control, simplifying the generation of motor commands. Studies have demonstrated that various motor behaviors can be reconstructed  with high fidelity using a basic set of coordinated muscle activity patterns endowed with different weights \cite{d2003combinations, d2006control}. We conceptualize muscle synergies as representations of actuators' control strategy. According to our hypothesis, these representations should depend on the physical structure of the controlled system, as this structure determines the characteristics of actuators. These representations should also exhibit generalizability across diverse tasks and conditions, reflecting commonalities within motor systems. Furthermore, we assume that these synergies can be adaptively fine-tuned to suit specific demands of each movement and state, as the actual working conditions of actuators are not inherently identical when performing different movements. We explore to discover synergistic representations from dynamical structures and embed them into deep RL methods to achieve efficiency and generalization in physiological motor control tasks.

In this work, we propose \algo, a deep reinforcement learning algorithm that is capable of generating interpretable synergistic representations of dynamical structures and performing task-specific, state-dependent adaptation to the representations. The generation process is driven by random perturbations. A stable and interpretable representation of dynamical synergies can be obtained. Embedding the representation into the learning process improve the efficiency of the agent when learning motor control policies demonstrated in Figure \ref{fig:motor_behaviors}. 

Our work mainly achieves following contributions: \textbf{(1)} We propose a method to generate synergistic representations from dynamical structure (for the first time), and a learning algorithm for state-dependent, task-specific adaptation in motion control tasks. \textbf{(2)} Experiments show that this representation generation method can obtain stable and interpretable representations of different overactuated systems. The representations can be generalized across different tasks. \textbf{(3)} We demonstrate that our representation-embedded learning algorithm can make the training process more efficient. \textbf{(4)} Our algorithm achieves control of ultra-high-dimensional musculoskeletal models, while other algorithms fail. \algo\ advances simulation control approaches in biomechanics, neuroscience and motor control communities.


%% file: sections/relatedwork.tex
\section{Related Work}

\textbf{Over-redundant actuation control.} Overactuated systems, often observed in the motor control of vertebrates, such as musculoskeletal systems, present a challenge for controllers trained by RL algorithms. A low-dimensional representation can be used to evaluate the quality of the control \cite{sui2017quantifying}. A series of attempts aimed on tackling the problem of learning control policies for locomotion and manipulation tasks with musculoskeletal models \cite{kidzinski2020artificial, MyoChallenge2022, MyoChallenge2023}. Leading solutions of these challenges include heavy curriculum training approaches, with reward shaping or demonstration imitation \cite{song2021deep}. Recently, a hierarchical reinforcement learning algorithm combined with imitation learning was applied to a 346-muscle musculoskeletal model \cite{lee2019scalable}. Generative models like variational autoencoders were utilized to control this musculoskeletal model to generate diverse behaviors \cite{park2022generative, feng2023musclevae}. In addition to human models, a musculoskeletal model of ostrich was constructed using the MuJoCo physics engine and controlled by TD4 deep reinforcement learning algorithm \cite{la2021ostrichrl}. Recent works, such as DEP-RL \cite{schumacher2023deprl} and Lattice \cite{chiappa2023latent}, have shown that employing better exploration techniques in reinforcement learning can help address the problem. Multi-task learning method is used in dexterous physiological control on a human hand model \cite{caggiano2023myodex}. Bio-inspired approaches \cite{cheng2019motor, berg2023sar} have demonstrated their effectiveness in motion control by applying synergistic representations in distinct parts of musculoskeletal bodies.

\textbf{Synergies for motor control in neuroscience.} For typical redundant actuation systems in nature, the coordination of actuators in vertebrates' neuromusculoskeletal motor control is known as muscle synergies. This can be defined as the coordinated recruitment of groups of muscles in the spatial, temporal, or spatiotemporal domains \cite{zhao2022evaluation}. As a long-standing theory in neurophysiology, muscle synergies is widely considered a possible approach for the central nervous system to overcome the complexity of motor control by reducing the number of independent parameters to simplify the generation of motor commands \cite{grillner1985neurobiological, dominici2011locomotor}. Researchers can generate the representation of muscle synergies through multidimensional matrix factorization from animals including rat, frog and human \cite{tresch2000motor, d2003combinations, ivanenko2004five, ting2005limited}. In experiments where humans perform fast reaching movements, changes of the muscle contraction patterns among various conditions can be explained with a high degree of confidence by assigning a set of synergy coefficients  \cite{d2006control}. The methodology of learning lower-dimensional action representations is being studied in the field of general robotics as well \cite{zhou2021plas, allshire2021laser, aljalbout2023clas}.

Our algorithm uses DRL to train control policies on high-dimensional, overactuated systems. Compared to existing methods, our representation generation method avoids training cost in an interpretable way. To the best of our knowledge, our work is the first algorithm to discover synergistic representations from dynamical structures and successfully use them to solve the motor control problem of high-dimensional overactuated systems.

%% file: sections/dynamical_feature.tex
\section{System Dynamical Features}

In this study, we aim to generate synergistic representations of actuators based on the dynamical characteristics of overactuated systems. Overactuation is common in natural musculoskeletal systems controlled by multi-articulation and pull-only actuations, making their motion control much harder than conventional torque-controlled robots \cite{caggiano2023myodex, berg2023sar}. In this section, we will introduce the neuro-musculoskeletal control of a full-body model as an example and outline the problem formulation.

\subsection{Physiological Neuro-Musculoskeletal Control} \label{Model}

\begin{figure}
\centering    
\subfigure[] {
\label{fig:full_model}
\includegraphics[width=0.45\linewidth]{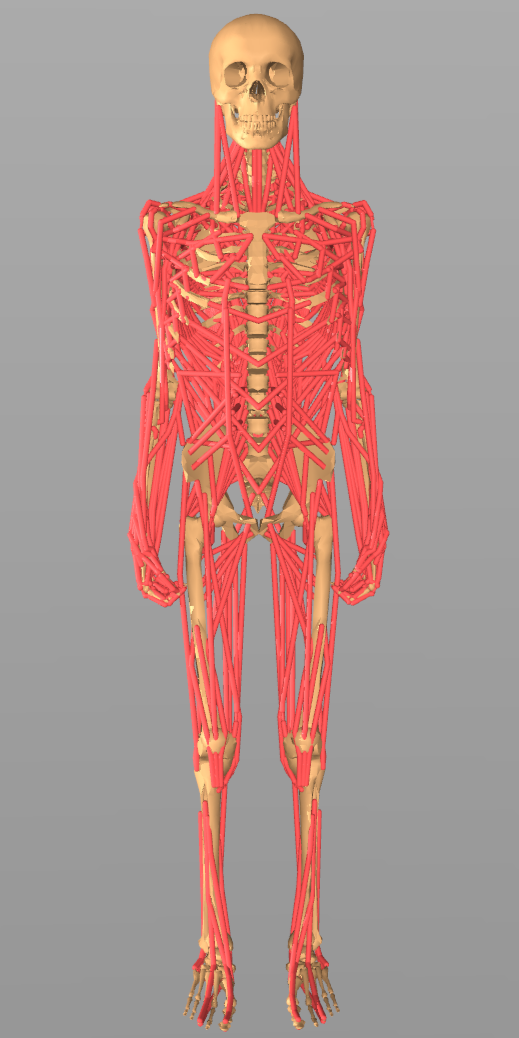} 
}     
\subfigure[] { 
\label{fig:arm_model}     
\includegraphics[width=0.45\linewidth]{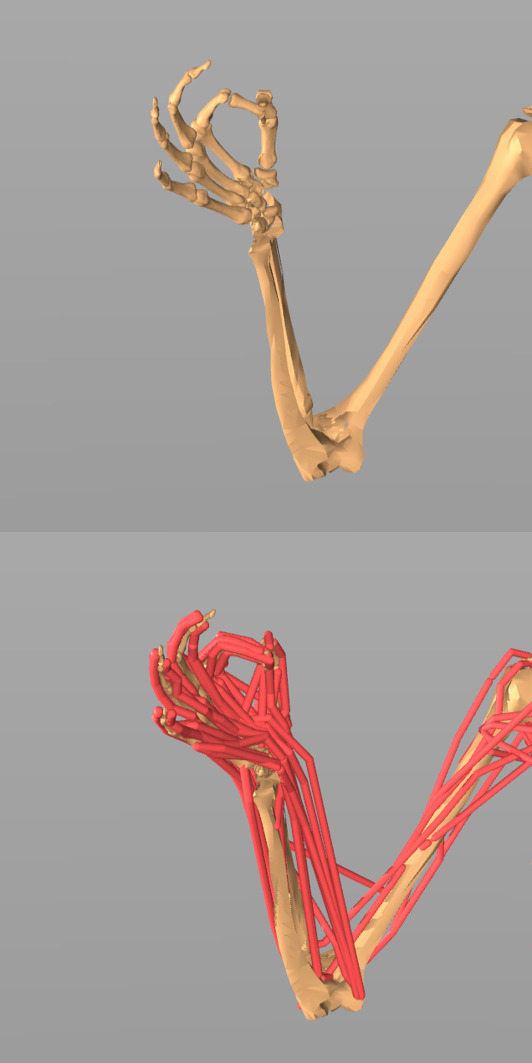} 
}    
\vskip -0.1in
\caption{\textbf{Musculoskeletal model.} (a) Full body model \model, red lines represent muscle-tendon units.  (b) Arm model with wrist and finger joints for dexterous manipulation. The top illustrates the the skeleton structure and the bottom illustrates muscles.} 
\label{fig:model}     
\end{figure}

\textbf{Full body musculoskeletal model.} As shown in Figure \ref{fig:full_model}, a full body musculoskeletal model \model~ \cite{zuo2023self} is used. The model has 90 rigid body segments, 206 joints, and 700 muscle-tendon units, and is implemented in the MuJoCo physics simulator \cite{todorov2012mujoco}. The base segment of the model is pelvis, which can translate and rotate in full degrees of freedom. The body parts can interact with the environment during simulation because the mesh files of their bones are used for  collision calculation. The dynamics of the human musculoskeletal system can be formulated with the Euler-Lagrangian equation as
\begin{equation}
    M(q)\Ddot{q} + c(q,\Dot{q}) = J_{m}^T f_m(act) + J_{c}^T f_{c} + \tau_{ext}
\end{equation}
where $q$ represents the generalized coordinates of joints, $M(q)$ represents the mass distribution matrix, and $c(q,\Dot{q})$ stands for Coriolis and gravitational forces applied to the generalized coordinates. $f_m(act)$ is the vector representing muscle forces generated by all muscle-tendon units, and is determined by muscle activations ($act$). $f_{c}$ is the constraint force and torque. $J_{m}$ and $J_{c}$ are Jacobian matrices that map force and torque to the space of generalized coordinates. $\tau_{ext}$ is external force and torque. The input control signal of muscle-tendon units is the neural excitations, which determine muscle activations. With the employment of the Hill-type muscle model \cite{zajac1989muscle}, the activation-contraction dynamics of muscles exhibit non-linearity and temporal delay, thereby posing challenges to neuromuscular control (see Appendix \ref{muscle_dynamics}). 

In Section \ref{Exp}, we apply \algo\  to several local models of human body (such as an arm model in Figure \ref{fig:arm_model}) and a model of ostrich \cite{la2021ostrichrl}. These local models of human body are part of the \model\ model, with slight differences in the implementation of simulation. The details of human local models and the ostrich model are presented in Section \ref{ExpSetup}.

\subsection{Problem Formulation}

A motor control task of musculoskeletal models and robots can be formulated as a Markov decision process, denoted by $\mathcal{M} = \langle \mathcal{S}, \mathcal{A}, r, p, \rho_0 , \gamma \rangle $, where $\mathcal{S} \subseteq \mathbb{R}^n$ represents the continuous space of all valid states, and $\mathcal{A} \subseteq \mathbb{R}^m$ represents the continuous space of all valid actions. $r : \mathcal{S} \times \mathcal{A} \to \mathbb{R}$ is the reward function. The state transfer probability function $ p(s'|s,a)$ describes the probability of an agent taking an action $a$ to transfer from the current state $s$ to the state $s'$. $\rho_0$ is the probability distribution of the initial state with $\sum_{s_0\in \mathcal{S}} \rho_0(s_0)=1$ and $\gamma \in [0, 1)$ is the discount factor. In the reinforcement learning paradigm, the goal of the agent is to optimize the policy parameter $\theta$ that maps from states to a probability distribution over actions $\pi_\theta : \mathcal{S} \to P( \mathcal{A})$. The policy seeks to maximize the discounted returns, $\pi^*_\theta(a|s)=\arg\max_\theta[\sum_{t=0}^{T} \gamma^t r(s_t,a_t)]$. The details of our action space and state space are described in Appendix \ref{env_detail}.

%% file: sections/dynamical_synergies.tex
\section{Dynamical Synergistic Representation} 

\begin{figure}[t]
  \begin{center}
  \includegraphics[width=1\linewidth]{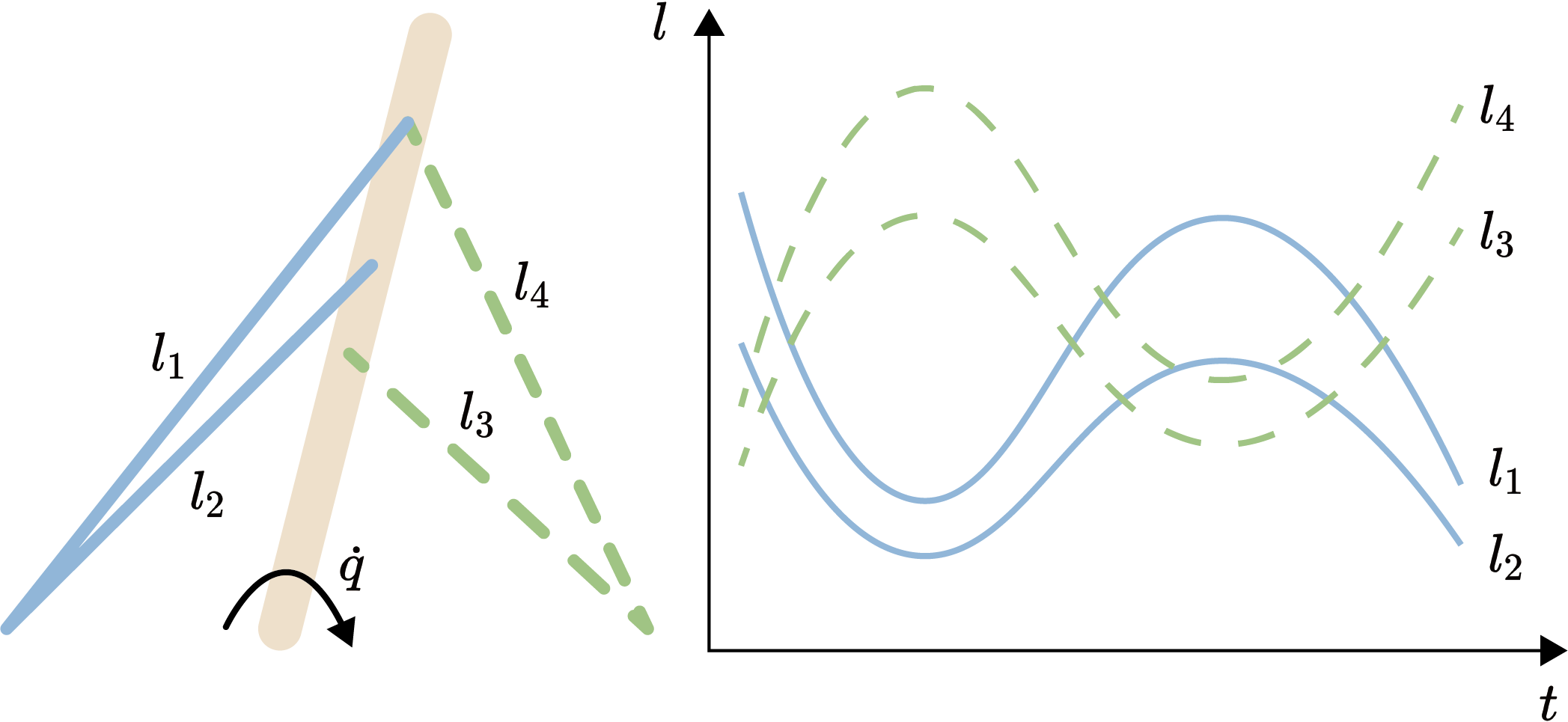}
  \vskip -0.1in
  \caption{\textbf{Motivation of \algo.} The brown link represents a robot arm (or bone), while the blue and green lines represent the cable actuators (or muscles). By randomly controlling the joint velocity, the lengths of the four actuators are demonstrated on the right. Actuators with similar functions are categorized into the same group due to similar structures, based on the correlation of length changes.}
  \label{fig:demo}
  \end{center}
  \vskip -0.1in
\end{figure}




As the action space enlarges, the sample efficiency of DRL algorithms sharply decreases. Researchers have explored various aspects of a typical example of these problems, human musculoskeletal system control, by means including refining exploration strategies \cite{schumacher2023deprl, chiappa2023latent} and the utilization of hierarchical learning approaches \cite{lee2019scalable}. Efforts has been made to learn synergistic action representations from trajectories in pre-training stage to expedite training, which is highly reliant on pre-training outcomes \cite{berg2023sar}. As shown in Figure \ref{fig:demo}, we observe that actuators with similar functions exhibit structural similarities. Hence, we employ a dynamics-based method which is able to rapidly generate interpretable synergistic representation. We then propose a novel algorithm to use these representations for further learning process.

\begin{algorithm2e}[t]
\SetAlgoLined
\DontPrintSemicolon
\LinesNumbered
\caption{
[\algo] Dynamical Synergistic Representation 
}

\label{alg:algo}

\KwIn{model $\mathcal{M}$, total trajectory steps $N_s$, control frequency $T$, control amplitude $A_c$, number of groups $N_g$}
\KwOut{grouping bins $G$}

Initialize trajectory buffer $\tau \leftarrow \emptyset$, $t \leftarrow 0$\;
\tcp{Random trajectory generation}
\While {$t \le N_s$}{
    \If{$t \bmod T = 0$}{
        Sample joint velocity $\dot q_t \sim \text{Unif}[-A_c, A_c]$\;
        Set model's joint velocity $\dot q \leftarrow \dot q_t$\;
    }
    Simulation step($\mathcal{M}$, zero action)\;
    $\tau \leftarrow \tau \bigcup l_t$ // Store the lengths of muscles \;
    $t \leftarrow t + 1$
}

\tcp{Grouping based on the correlation}
Calculate correlation matrix $R$ using $\tau$ with Equation \ref{eq:corr}\;
Grouping bins $G \leftarrow \text{K-Medoids}(1 - R)$
\end{algorithm2e}

\begin{figure*}[ht]
  \begin{center}
  \includegraphics[width=0.92\linewidth]{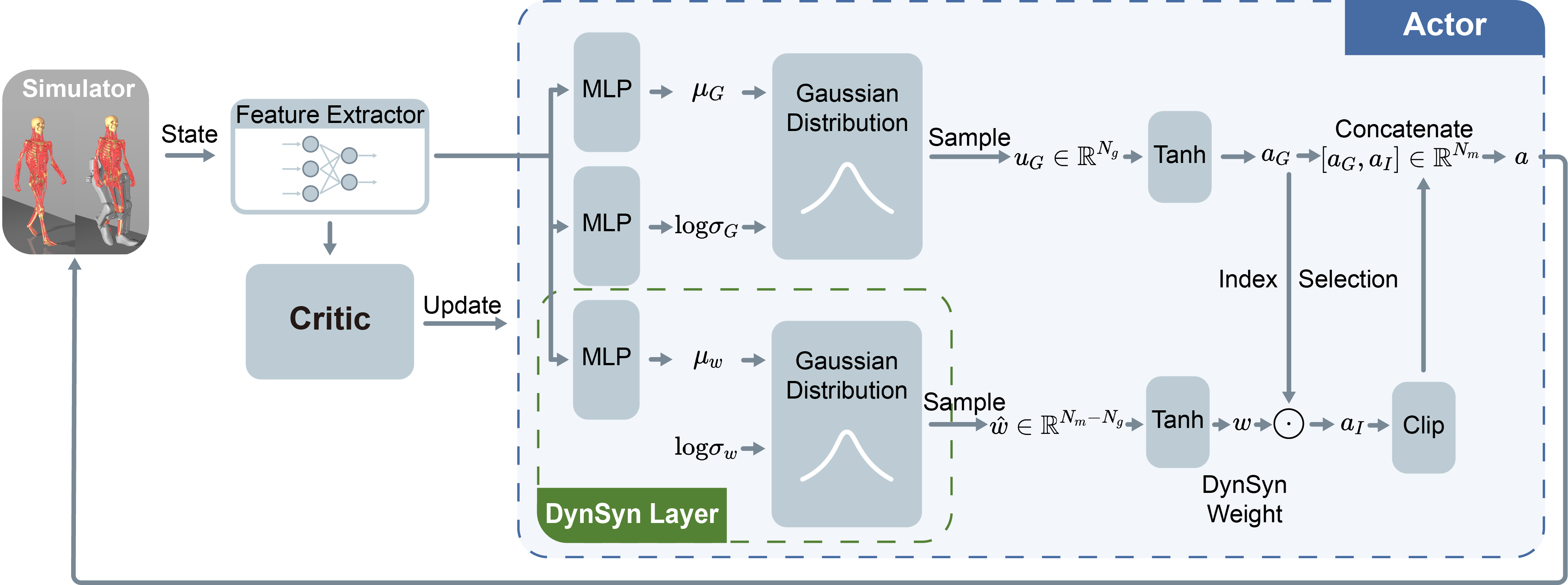}
  \vskip -0.1in
  \caption{\textbf{Overview of \algo.} The algorithm generates a unified action $a_G$ for each group of actuators, along with state-dependent correction weights $w$ for each actuator on top of the unified action $a_G$. Separate MLPs are used to generate the parameters of two Gaussian distributions. We then sample from the Gaussian distributions and pass them through a squashing function Tanh to obtain $a_G$ and $w$.}
  \label{fig:algo}
  \end{center}
\end{figure*}

\subsection{Representation Generation} \label{representation_generation}
As illustrated in Algorithm \ref{alg:algo}, We employ a similarity-based grouping method for the dynamical synergistic representation generation. Firstly, we generate muscle length trajectories of length $N_s$ through applying random control to the joint space of a musculoskeletal model $\mathcal{M}$. Here, the control signal is the joint velocity $\dot q$ of the musculoskeletal model, and this signal is sampled randomly from a uniform distribution $\text{Unif}[-A_c, A_c]$ every $T$ time intervals. Upon obtaining the muscle length trajectory $\tau \in \mathbb{R}^{N_s \times N_m}$, we calculate the correlation between length changes for each pair of muscles according to Equation \ref{eq:corr}. $N_m$ is the dimension of actions (the number of muscles). Based on the correlation matrix $R \in \mathbb{R}^{N_m \times N_m}$, and a predetermined number of groups $N_g$, we employ the K-Medoids \cite{park2009simple} clustering algorithm to generate the closest clustering results, forming grouping bins $G$.
\begin{equation}
\label{eq:corr}
    R_{i,j} = \frac{1}{N} \sum_{k=0}^{N-1} S_c(\tau^i_{[\frac{N_s}{N}k: \frac{N_s}{N}(k+1)]}, \tau^j_{[\frac{N_s}{N}k: \frac{N_s}{N}(k+1)]}) 
\end{equation}
The correlation matrix is calculated by Equation \ref{eq:corr}. We divide the trajectories into $N$ segments with respect to time and then average the similarity of each segmented trajectory. $N$ represents the number of segmented trajectories, $S_c$ is the cosine similarity, and $i$, $j$ correspond to the $i$ th and $j$ th muscles, respectively. Subscript $[t_1:t_2]$ represents the trajectory from time $t_1$ to time $t_2$.

\subsection{State-dependent Representation} \label{State-dependent Representation}
Using the above algorithm, functionally similar actuators will be categorized into a group, and assigned with same actions. This prevents DRL algorithms from assigning opposite actions to functionally similar actuators during the learning process, thereby enhancing effective exploration in high-dimensional action spaces. However, merely assigning same actions to all actuators within a group may result in unnatural movements. Therefore, we propose the algorithm illustrated in Figure \ref{fig:algo}. While the actuators within a group perform shared actions, state-dependent correction weights are produced for each actuator to facilitate fine-tuned adjustments.

Based on the SAC algorithm \cite{haarnoja2018soft}, our algorithm generates a unified action $a_G$ for each group, along with state-dependent correction weights $w$ for each actuator on top of the unified action $a_G$. $a_G$ and $w$ are written as
\begin{align}
    a_G =& \text{ tanh}(u_G), u_G \sim \mathcal{N}(\mu_G, \sigma_G) \\
    w =& \text{ tanh}(\hat w), \hat w \sim \mathcal{N}(\mu_w, \sigma_w),
\end{align}
where $\mu_G$, $\sigma_G$ represent the mean and variance of the unified actions, $\mu_w$ is the mean of state-dependent
correction weights and $\sigma_w$ is the state-independent variance of the weights. By default, the first actuator in each group is assumed to have a correction weight of 1, and $N_m - N_g$  correction weights are to be determined. The final action is computed using the following equations:
\begin{align}
        a_I =& \text{ IS}(a_G) \odot \text{clip}(\kappa w, -c, c) \\
        c =& \min( \max (k_D t + a_D, 0), \kappa) \\
        a =& \text{ clip}([a_G, a_I], -1, 1),
\end{align}
where $\text{clip}(x, l, h)$ is a function that restricts the value of $x$ to the interval $[l, h]$. The Index Selection function ($\text{IS}$ function),  selects corresponding unified actions $a_G$ according to indices determined by grouping results.  $\odot$ represents element-wise multiplication, and $[\cdot, \cdot]$ denotes concatenation. $a_I$ is the individual action. $k_D$, $a_D$ and $\kappa$ are the hyperparameters of the weight boundary, which relaxes gradually as the training timesteps $t$ increase. During training, the state-dependent adaptation of representations will start at $a_D$ considering the stability of learning. Finally, the policy $\pi$ will be updated according to the following formula:
\begin{equation}
    \pi^* = \arg \max_\pi \underset{\tau \sim \pi}{\text{E}}[\sum_{t=0}^{\infty} \gamma^t(R(s_t,a_t) + \alpha H(\pi(a_G|s_t)))]
\end{equation}

%% file: sections/experiments.tex
\section{Experiments} \label{Exp}

We demonstrate our method's efficiency during learning and its generalization ability in overactuated motor control benchmarks built in MuJoCo. In this part, we will introduce the benchmarks, the learning process of \algo\ and the details of baselines.

\begin{figure*}[htbp]
  \begin{center}
  \includegraphics[width=1\linewidth]{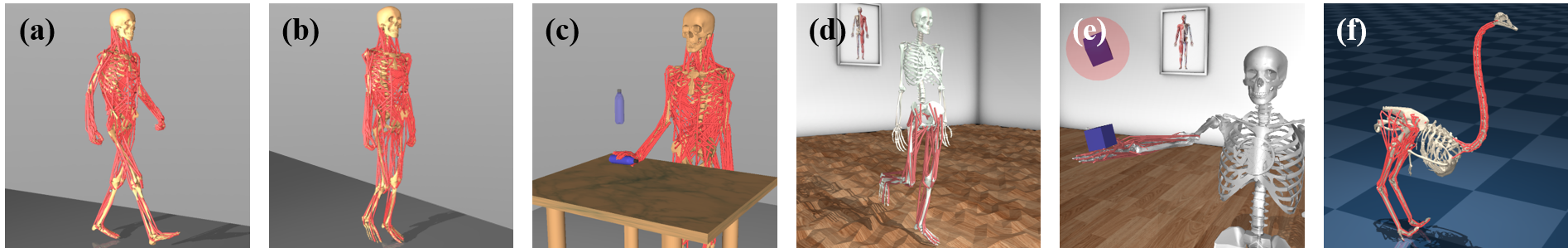}
  \vskip -0.1in
  \caption{\textbf{Experiment environments.} (a) \model-Gait ($a\in \mathbb{R}^{700}$). (b) Legs-Walk ($a\in \mathbb{R}^{100}$). (c) Arm-Locate ($a\in \mathbb{R}^{81}$). (d) MyoLegs-Walk with rough terrain ($a\in \mathbb{R}^{80}$). (e) MyoHand-Reorient100 ($a\in \mathbb{R}^{39}$). (f) Ostrich-Run ($a\in \mathbb{R}^{120}$). The semitransparent objects in manipulation environments are the target indicators.}
  \label{fig:envs}
  \end{center}
\end{figure*}

\subsection{Environments} \label{ExpSetup}

We create reinforcement learning environments of various models and tasks to test our algorithm. Additionally, two tasks from MyoSuite \cite{MyoSuite2022} are taken into account. We use \textbf{Model-Task} pair to label the environments, as shown in Figure \ref{fig:envs}. A complete description including the action space, the state space and the reward function of each environment is given in Appendix \ref{env_detail}.

\textbf{Human Motion Imitation}: In \textbf{FullBody-Gait}, we expect the full body \model\ model with 206 joints actuated by 700 muscles (described in Section \ref{Model}) to mimic a motion-capture walking trajectory. During training, the model may be initialized at any time step throughout a trajectory cycle. 

\textbf{Human Locomotion}: In \textbf{Legs-Walk}, a 20-DoF \textit{Legs} model actuated by 100 muscles is used. In \textbf{MyoLegs-Walk}, the \textit{MyoLegs} model in MyoSuite with 20 DoF and 80 muscles is used. Both models are expected to walk forward robustly, driven by biomechanically inspired reward functions. 

\textbf{Human Manipulation}: In \textbf{Arm-Locate}, an \textit{Arm} model of 28 DoF and 81 muscles, with wrist and fingers is used. The agent is trained to learn to grasp a bottle, relocate it to the random target position and orientation. In \textbf{MyoHand-Reorient100}, the \textit{MyoHand} model in MyoSuite with 23 DoF and 39 muscles needs to rotate a set of 4 objects, each with 25 different geometries, to predetermined orientations. 

\textbf{Animal Locomotion}: In \textbf{Ostrich-Run}, an \textit{Ostrich} model \cite{la2021ostrichrl} with 50 joints actuated by 120 muscles is used. The model is trained to run horizontally as fast as possible by rewarding its velocity.

\textbf{Generation Tasks}: We test various terrain conditions (\textbf{MyoLegs-Walk-Terrain}) and different walking targets (\textbf{Legs-Walk-Fast, Legs-Walk-Diagnal}) to demonstrate the generalization capability of the dynamical synergistic representation across various physical conditions, as well as the robustness of generated motor behaviors.


The \textit{Legs} model and the \textit{Arm} model are obtained by removing irrelevant degrees of freedom from the full body model. For the locomotion task and the manipulation task, two environments are tested for each task to demonstrate the robustness of the algorithm, given that there are several variations between the models and the environments.

\subsection{Learning Dynamical Synergistic Representation} \label{learning_synergy}

Before training, dynamical synergistic representations (i.e. the grouping of actuators) are generated for each model. We impose random control on joint velocities for $5e5$ simulation frames to collect sequences of actuators' features. For muscle-tendon units, the collected feature is length. The grouping of actuators is then obtained according to Section \ref{representation_generation}. We choose an appropriate number of groups where the difference between maximum and minimum distances among cluster centers are large enough. This process is further detailed in Appendix \ref{group_num}. To demonstrate that the representation generated by our method can stably capture the dynamical features of the models, we repeat the generation for 10 times on each model and calculate the mean value and variance. The same representation of a single model are retained in tasks with changed conditions to verify the generalization ability of the algorithm. Furthermore, a series of ablation experiments are presented to prove that our choice of the number of groups is reasonable and helpful for the learning of motor control (see Section \ref{Syn_Gen}).

\subsection{Baselines}

We compared \algo\ with current DRL methods in overactuated systems: SAC \cite{haarnoja2018soft}, SAR \cite{berg2023sar} and DEP-RL \cite{schumacher2023deprl}. SAR collects low-dimensional representations from a pre-training collection stage over $M$ time steps and its training stage is over another $N$ time steps. Other methods are directly trained over $M+N$ time steps. It should be noted that DEP-RL is an exploration method which can be integrated into our algorithm. \algo\ are based on the RL library Stable-Baselines 3 \cite{raffin2019stable}. Hyperparameters and implementation details in the experiments are summarized in Appendix \ref{algo_detail}. All results are averaged across 5 random seeds.

%% file: sections/results.tex
\section{Results and Analysis}

\begin{figure}[ht]
  \begin{center}
  \includegraphics[width=1\linewidth]{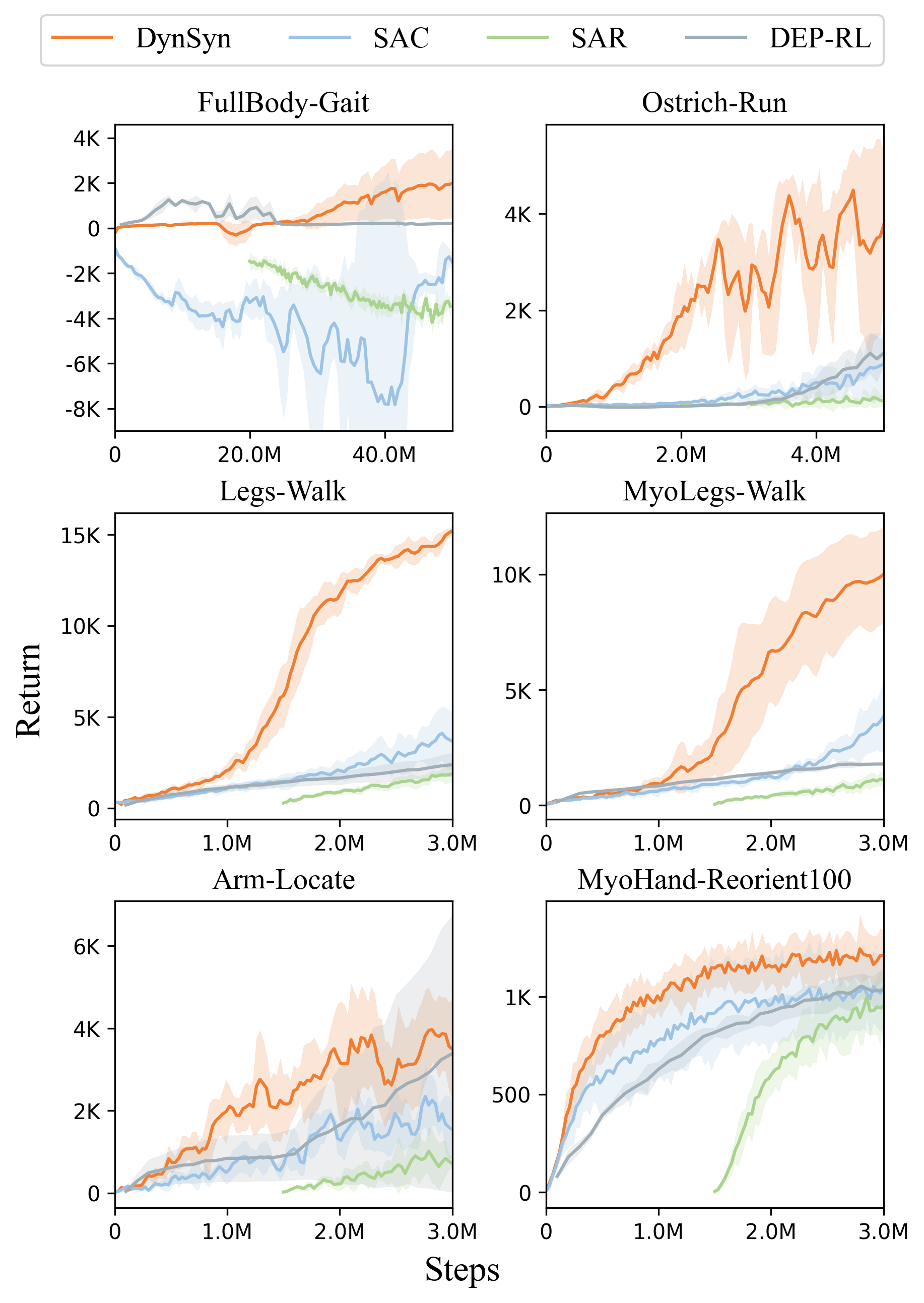}
  \vskip -0.1in
  \caption{\textbf{Standard experimental results.} Learning curves in the experimental environments. Mean $\pm$ SD across 5 random seeds for all the environments. SAR are depicted to begin at $M$ timesteps in order to account for the steps of pre-training. The return of baselines decreases as the number of action dimensions increases, while \algo\ is the only algorithm that performs well even in a very high-dimensional action space of 700 dimensions in the \textbf{FullBody-Gait} environment. }
  \label{fig:result_1}
  \end{center}
  \vskip -0.1in
\end{figure}

\begin{figure}[ht]
  \begin{center}
  \includegraphics[width=1\linewidth]{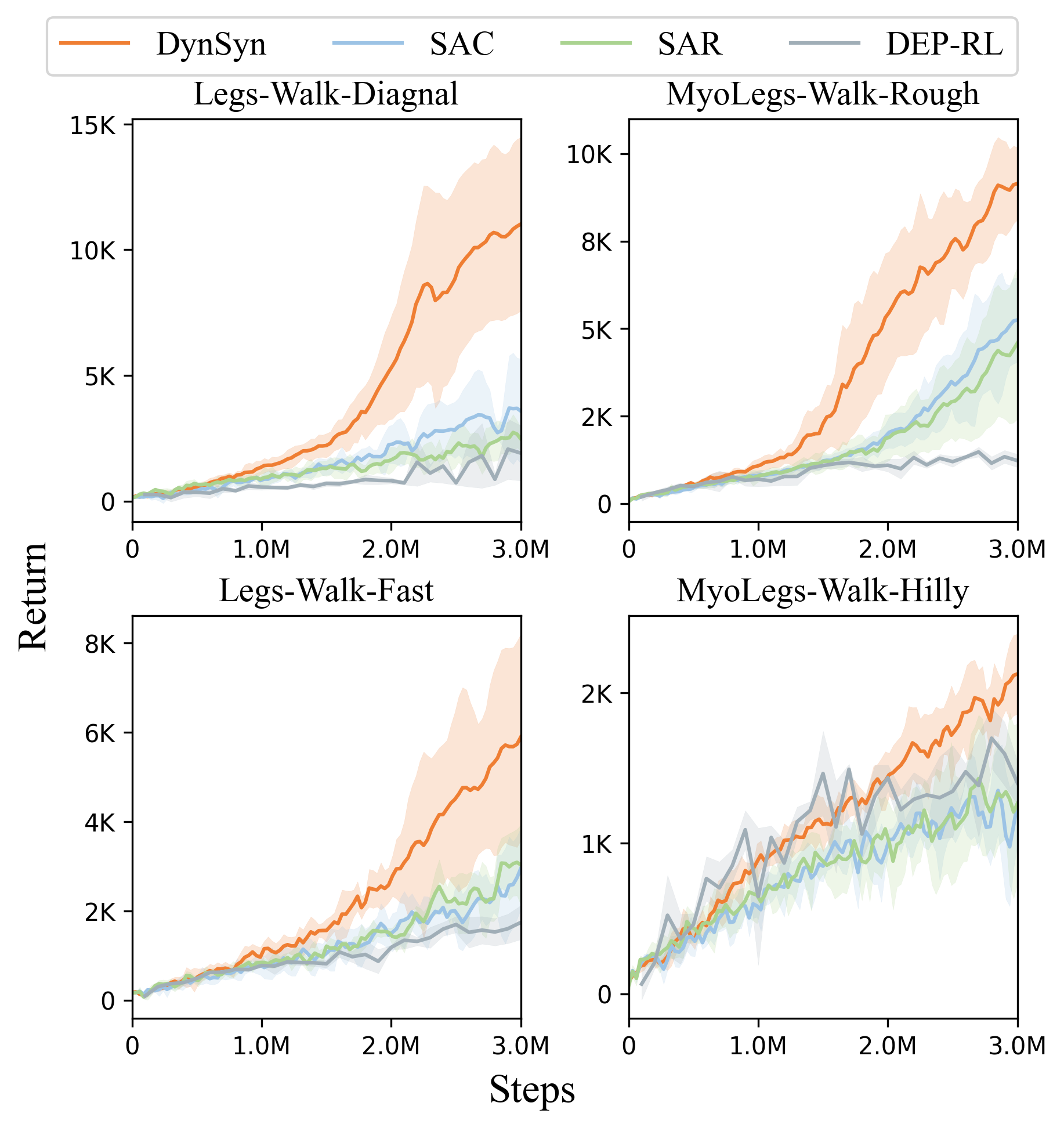}
  \vskip -0.1in
  \caption{\textbf{Generalization experimental results.} Learning curves in the generalization environments in which the physical situations or targets are changed. Mean $\pm$ SD across 5 random seeds for all the environments. Here, the $M$ of SAR is $0$ because the representations are generated from standard environments.}
  \label{fig:result_2}
  \end{center}
  \vskip -0.1in
\end{figure}


In this section, we present the experimental results, demonstrating that \algo\ effectively facilitates motor control across various tasks involving different models, exhibiting state-of-the-art sample efficiency and high stability. Additionally, we illustrate that the dynamical synergistic representations extracted from models exhibit good performance in terms of convergence and interpretability. This allows the model to leverage the representations in learning motor control across diverse tasks, even under varying conditions such as terrain and training targets.

\subsection{Efficient Learning}

Figure \ref{fig:result_1} illustrates that \algo\ achieves higher returns in fewer training steps across all standard experimental environments. This implies that \algo\ efficiently produces robust motor control (refer to Figure \ref{fig:motor_behaviors}) in various overactuated models and motor tasks. Notably, the performance of baseline agents significantly deteriorates as the number of action dimensions increases, whereas \algo\ performs well even in a very high-dimensional action space of 700 dimensions.


\subsection{Synergies Generalization} \label{Syn_Gen}

\begin{figure*}[htbp]
\centering    
\subfigure[] {
\label{fig:group_matrix}
\includegraphics[width=0.43\linewidth]{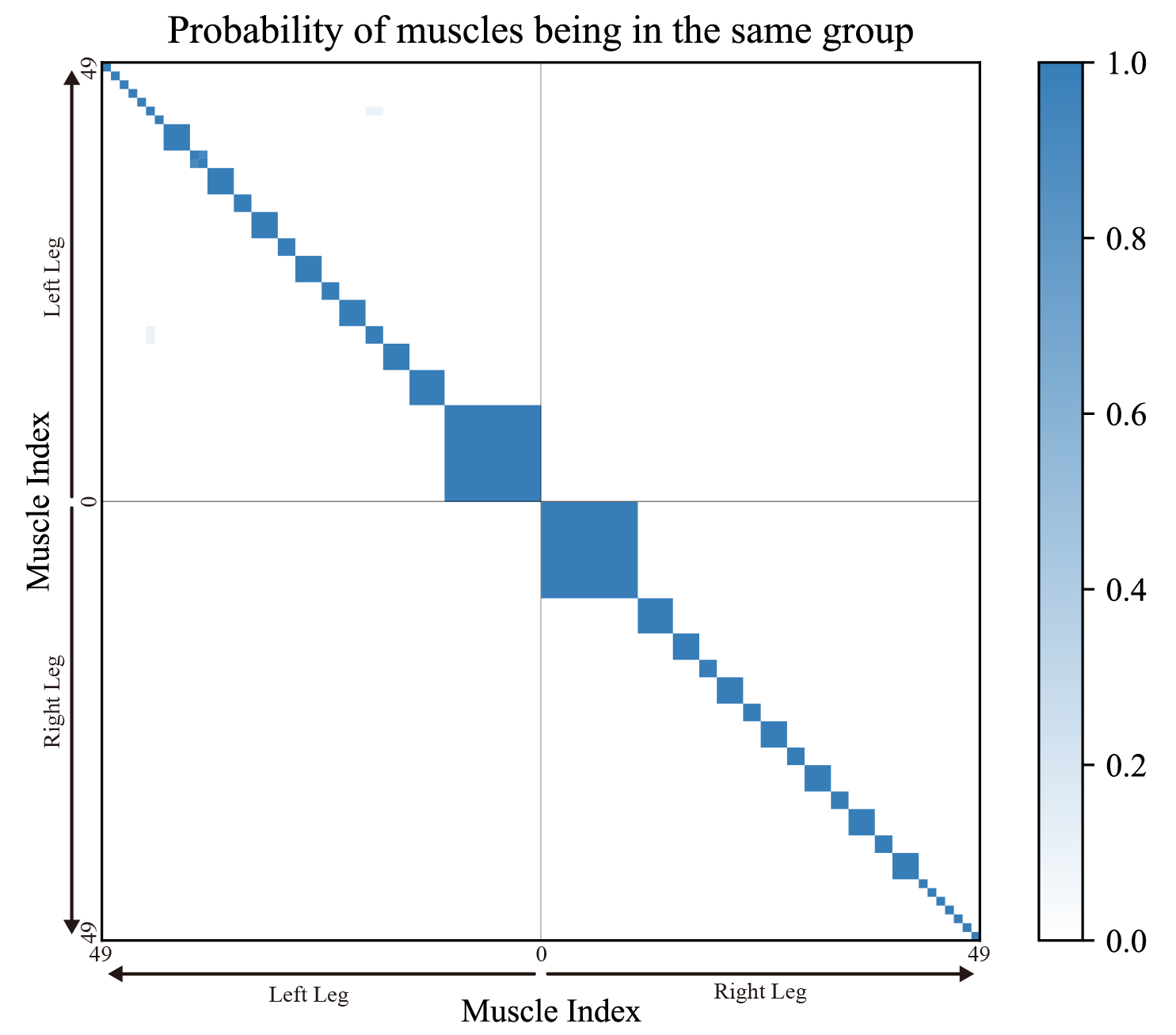} 
}     
\subfigure[] { 
\label{fig:group_visual}     
\includegraphics[width=0.35\linewidth]{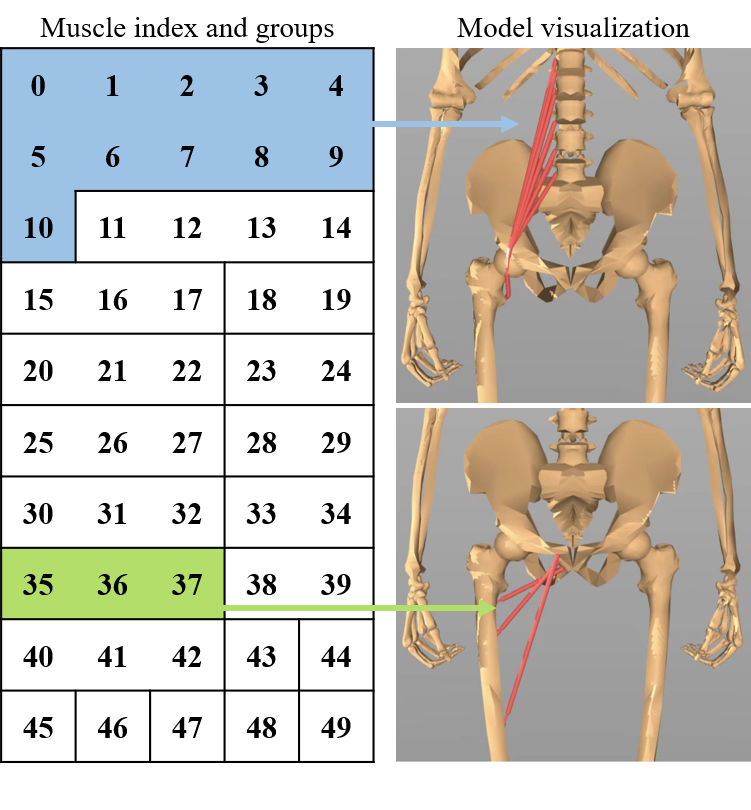} 
}    
\vskip -0.1in
\caption{\textbf{Muscle grouping of \textit{Legs} model.} (a) Grouping matrix $P_{n}$, where $p_{ij}$ is the probability that muscle $i$ and muscle $j$ are in the same group (averaged over 10 seeds). For the \textit{Legs} model where $n=100$, the first 50 muscle indices represent muscles of the right leg, and the other 50 indices represent the left leg. The model is bilaterally symmetrical, and the muscle indices are reordered for better visualization.  (b) Visualization of muscle groupings for the right leg. Each number represents a muscle, and different groups are delineated by borders.}     
\vskip -0.1in
\label{fig:group_result}
\end{figure*}

When the same representations are applied to tasks with additional environmental conditions or changed targets, such as rugged terrains and walking direction, \algo\ maintains good performance (see Figure \ref{fig:result_2}). This suggests that the generated synergistic representations of the same model can effectively generalize across different tasks. Figure \ref{fig:group_matrix} shows the average results of 10 preliminary group extractions for the \textit{Legs} model, showing a high probability of obtaining the same grouping result (close to 1) and furthermore, the stability of the extraction process.

\begin{figure}[h]
  \begin{center}
    \includegraphics[width=1\linewidth]{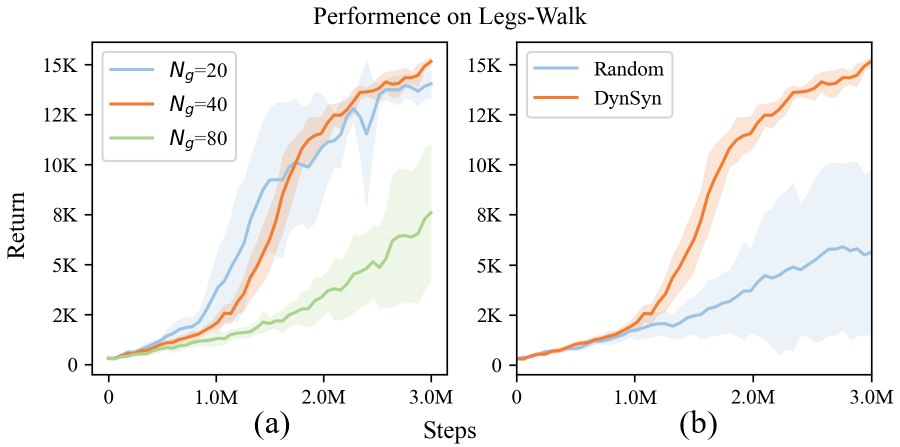}
    \vskip -0.1in
  \caption{\textbf{Ablation study results.} Learning curves on the \textbf{Legs-Walk} environment. Mean $\pm$ SD across 5 random seeds. (a) Performance variation with different numbers of clusters, we apply $N_g=40$ in our final control tasks. (b) Performance comparison between randomly generating 40 clusters and our method generating the dynamical synergistic representation of 40 clusters. }
  \label{fig:ablation}
  \vskip -0.2in
  \end{center}
\end{figure}

For ablation study, we utilize a random grouping approach to create 40 clusters for comparison. As shown in Figure \ref{fig:ablation}, our method consistently yields performance improvements. However, randomly generated representations outperform the SAC algorithm, possibly due to the influence of our state-dependent algorithm. The high standard deviation of the learning curve of randomly generated representation shows a decrease in stability. We also attempt to generate representations for varying numbers of clusters. The results demonstrate that our cluster number selection scheme ensures performance stability with a lower standard deviation of the learning curve (see Figure \ref{fig:ablation}). Each grouping mode is tried with 5 different random seeds.

\subsection{Physiological Analysis}

In accordance with the left-right symmetry in \textit{Legs} modeling, 50 indices, from 0 to 49, are assigned to muscles of each leg symmetrically. In the grouping matrix in Figure \ref{fig:group_matrix}, the upper left quadrant signifies the correlation between left-leg muscles, and the lower right quadrant represents the correlation between right-leg muscles. The remaining part of Figure \ref{fig:group_matrix} depicts correlation near 0 between pairs of muscles from both legs. Notably, our representation generation method identifies inherent symmetries in the musculoskeletal model. Only groupings within the same leg are observed. Furthermore, the groupings of muscles from the left and right legs are symmetrical. For improved visualization, the muscle grouping result of the right leg is expanded and depicted in Figure \ref{fig:group_visual}, detailing two representative muscle grouping examples in the musculoskeletal model (i.e., Psoas Major and Thigh Adductors). From a biomechanical perspective, this is evident that muscles grouped together exhibit similar effects, highlighting our method's ability to capture fundamental dynamical characteristics in the system.

%% file: sections/conclusion.tex
\section{Conclusion and Discussion}

We introduce \algo, a deep RL method that generates synergistic representations of actuators from dynamical structures of overactuated systems and make task-specific, state-dependent adaptation to the representations, thereby expediting and stabilizing motor control learning. Applying \algo\ to musculoskeletal locomotion and manipulation tasks, we demonstrate its superior learning efficiency compared to all baselines. Additionally, we illustrate the robust generalization ability of the extracted synergistic representations across various motor tasks with the same model. In conclusion, our work offers an efficient, generalizable, and interpretable approach to controlling high-dimensional redundant actuation systems. The generation method of synergistic representations can help deepen the understanding of motor intelligence. This research aims to facilitate the training of motor control policies for use in artificial intelligence, robotics and medicine, contributing to the development of a versatile embodied agent.

Despite the promising outcomes, distinctions persist between real embodied motor intelligence and the musculoskeletal model simulation employed in our study. For instance, current simulation methods primarily leverage proprioception (joint position and velocity), whereas in the real world, an animal receives additional sensory inputs, including vision and touch \cite{patla1997understanding, patla1998human, jeka2000multisensory}. To enhance customization for specific applications, further work on biomechanically realistic simulations is essential. Other significant limitations include the multiple potential solutions in overactuated systems, and our method can only generate one of the numerous high-dimensional combinations to control the system. Future research may need to consider establishing a solution space of control patterns.

%% file: sections/impact.tex
\section*{Impact Statement}

This paper presents work whose goal is to advance the field of Machine Learning, especially Embodied Intelligence. There are many potential societal consequences of our work, none which we feel must be specifically highlighted here.

%% file: sections/appendix.tex
\appendix

\section{Neuro-muscle dynamics} \label{muscle_dynamics}

The input control signal of muscle-tendon units is the neural excitation $ctrl$, and the muscle activation $act$ is calculated by a first-order nonlinear filter as follow:
\begin{equation}
    \frac{\partial act}{\partial t} = \frac{ctrl-act}{\tau(ctrl,act)}, \tau(ctrl,act) = \begin{cases}\tau_{act}(0.5+1.5act) & ctrl>act\\

    \tau_{deact} / ({0.5+1.5act}) & ctrl\leq act\end{cases},
    \label{equ:muscle_dynamic}
\end{equation}
$(\tau_{act}, \tau_{deact})$ is a time constant to activate or deactivate latency of defaults $(10ms, 40ms)$. The force produced by a single muscle-tendon unit can be formulated as 
\begin{equation}
f_m(act)=f_{max}\cdot [F_{l}(l_m)\cdot F_v(v_m)\cdot act + F_p(l_m)],
\end{equation}
where $f_{max}$ stands for the maximum isometric muscle force and $act, l_m, v_m$ respectively stand for the activation, normalized length and normalized velocity of the muscle. $F_l$ and $F_v$ represent force-length and force-velocity functions fitted using data from biomechanical experiments \cite{millard2013flexing}. 

\section{Environment Details} \label{env_detail}

For all environments, the simulation time step is 0.01s. The action space consists of muscle excitations $ctrl$ (i.e. motor neuron signals). The dimensions of action and state spaces, number of joints and episode length of all the environments are summarized in Table \ref{env_tab}. Task and reward parameters are summarized in Table \ref{env_para}.

\textbf{FullBody-Gait} \quad We expect the full body \model\ model with 206 joints actuated by 700 muscles (Section \ref{Model}) to mimic a motion-capture walking trajectory. During training, the model may be initialized at any time step throughout a trajectory cycle. The state space consists of simulation time, joint positions, joint velocities, muscle forces, muscle lengths, muscle velocities, muscle activation and reference joint positions. The reward function is:
\begin{align}
    R =& w_v R_v + w_q R_q + w_h \mathbb{I}_{\text{alive}} \\
    R_v =& \exp(-(v_x^{com} - v_x^t)^2) + \exp(-(v_y^{com} - v_y^t)^2) \\
    R_q =& -|| q - q^r ||_2
\end{align}
where $q$ is actual joint positions, $q^r$ is the reference joint positions, $\left \{v_x^{com}, v_y^{com}\right \}$ is the velocity of the center of mass and $\left \{v_x^{t}, v_y^{t}\right \}$ is the desired velocity. $w_v$, $w_{q}$ and $w_h$ are the weights. 

\textbf{MyoLegs-Walk} \quad The \textit{MyoLegs} model in MyoSuite with 20 DoF and 80 muscles is used. The model is expected to walk forward robustly, driven by biomechanically inspired reward functions:
\begin{equation}
    R = w_vR_v-w_cR_c+w_rR_r+w_jR_j-w_aR_a-w_dR_d
\end{equation}
$R_d= \{falled\}$, imposes a penalty when the model falls. The weights $w_v$, $w_c$, $w_r$, $w_j$, $w_a$, and $w_d$ determine the importance of each reward term. The other terms are as follow:
\begin{equation}
    R_v = \exp(-\sqrt{v_x^r-v_x}) + \exp(-\sqrt{v_y^r-v_y})
\end{equation}
$v^r$ and $v$ represent the desired and actual velocity of the center of mass. $R_v$ represents the velocity reward.
\begin{equation}
    R_c = ||[0.8\cos(\phi\times2\pi + \pi), 0.8\cos(\phi\times2\pi)] - [q_{rhip}, q_{lhip}]||
\end{equation}
$\phi$ is the phase percentage of the pre-define gait period. $q_{rhip}$ and $q_{lhip}$ are the hip flexion angle of both legs. $R_c$ encourages rhythmic hip movements.
\begin{equation}
    R_r = \exp(-5||(q_{pelvis}-q_{pelvis}^{init})||)
\end{equation}
$q_{pelvis}$ and $q_{pelvis}^{init}$ are the quaternions of pelvis and its initial value when reset. $R_r$ encourages the model to follow a predetermined rotation pattern. 
\begin{equation}
    R_j = \exp(-5\sum^N_{i=1}|q_i|/N)
\end{equation}
In the environment, $N=4$ and $q_i$ are the hip abduction and rotation angles of both legs. $R_j$ penalizes deviations from desired joint angles.
\begin{equation}
    R_a = ||act||/N_a
\end{equation}
$act$ is the muscle activation vector, $N_a$ is the number of muscles, and $R_a$ promotes efficient actuator usage by computing the norm of the action divided by the number of actuators.

\begin{figure*}[htbp]
\centering
\subfigure[] {
\label{fig:FlatTerrain}
\includegraphics[width=0.2\columnwidth]{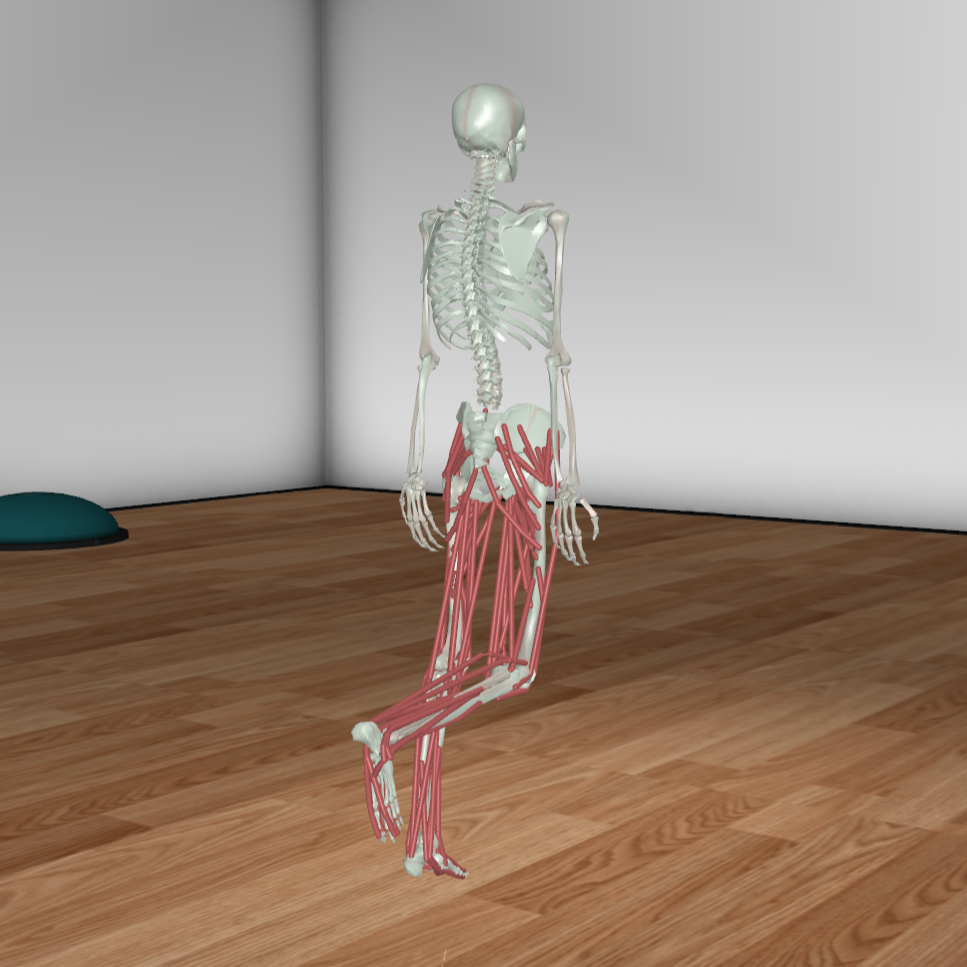}  
} 
\subfigure[] {
\label{fig:HillyTerrain}
\includegraphics[width=0.2\columnwidth]{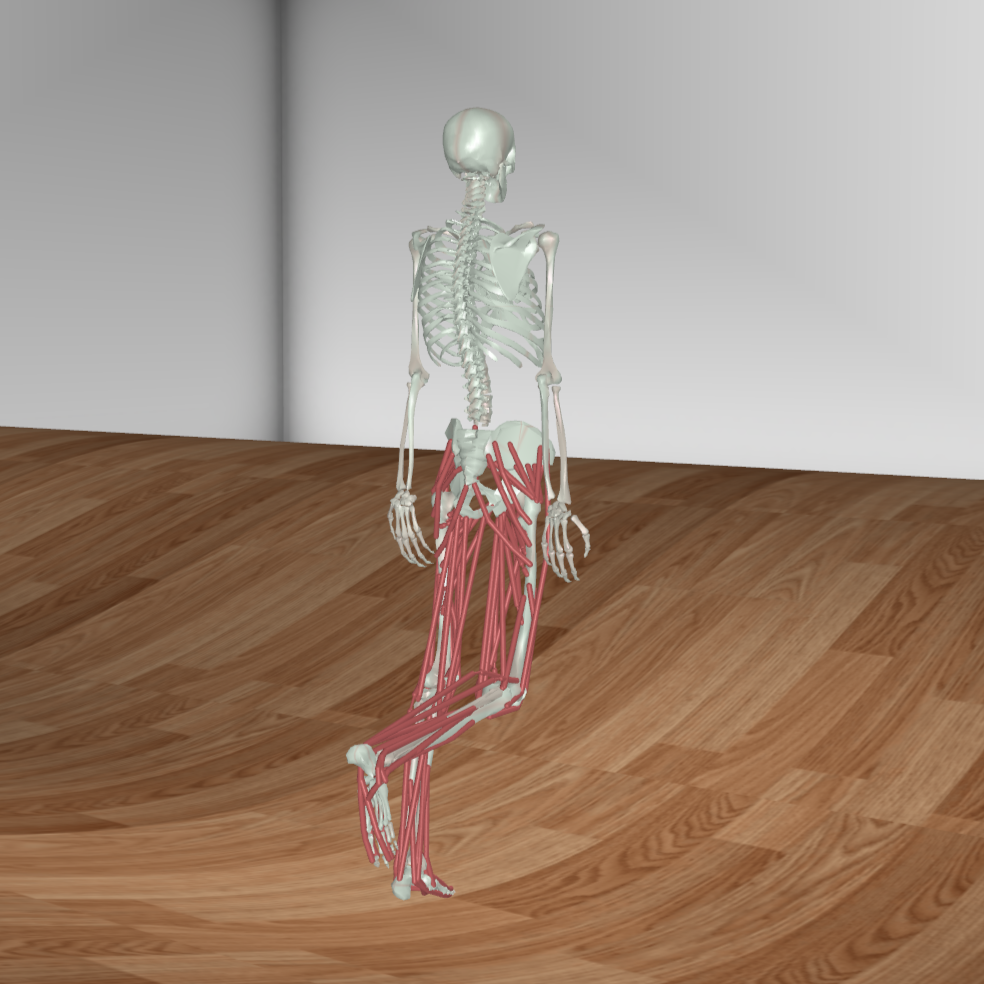}   
}
\subfigure[] {
\label{fig:RoughTerrain}
\includegraphics[width=0.2\columnwidth]{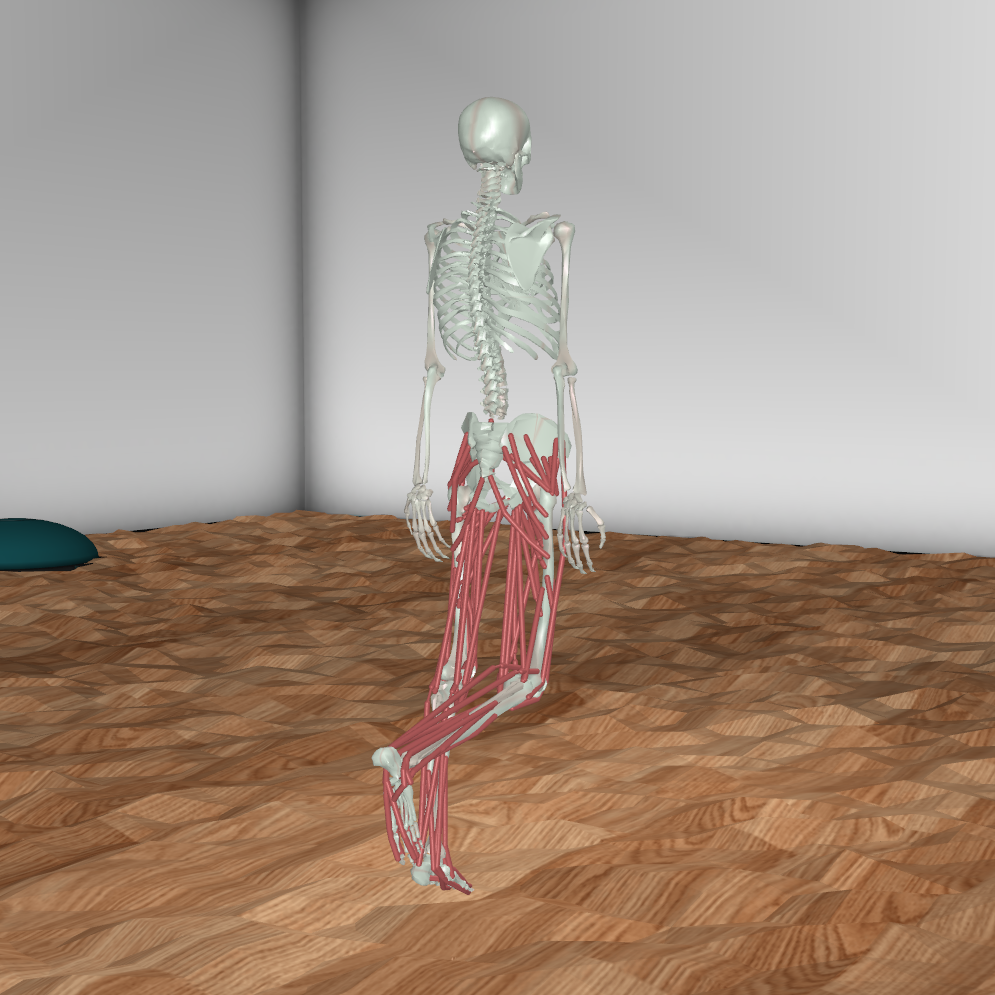}   
}
\caption{\textbf{MyoLegs-Walk terrains.} (a) Flat terrain. (b) Hilly terrain. (c) Rough terrain.}     
\label{fig:MyoLegsWalkTerrain}     
\end{figure*}

The state space consists of simulation time, joint positions (except $x$ and $y$ positions for the base segment), joint velocities, muscle forces, muscle lengths, muscle velocities and muscle activations. The task-specific observations include time phase percentage of the gait, velocity and height of the center of mass, torso angle, the height of the feet and their positions relative to the pelvis. The diverse terrain conditions including slopes and rough ground can be added to the task (see Figure \ref{fig:MyoLegsWalkTerrain}).

\textbf{Legs-Walk} \quad A 20-DoF \textit{Legs} model actuated by 100 muscles is used. The observations and reward function are the same as \textbf{MyoLegs-Walk} environment except that there is no $R_{a}$ term. In addition, we apply diverse walking speed targets in the simulation. 

\textbf{Arm-Locate} \quad An \textit{Arm} model of 28 DoF and 81 muscles with wrist and fingers is used. The agent is trained to learn to grasp a bottle, relocate it to the target position and reorient it to the target orientation. The position and orientation of target are randomized when reset. The reward function is:
\begin{equation}
    R = w_pR_p \times w_oR_o+w_rR_r-w_aR_a
\end{equation}
$R_a = ||act||/N_a$, promotes efficient actuator usage. We use multiplication to enforce the relocation and reorientation simultaneously. The weights $w_p$, $w_o$, $w_r$ and $w_a$ determine the importance of each reward term. The other terms are as follow:
\begin{equation}
    R_p = \exp(-10\sqrt{||p_{target}-p_{object}||})
\end{equation}
$p_{target}$ and $p_{object}$ represent the positions of the target and the object, respectively. $R_p$ represents the reward for relocation.
\begin{equation}
    R_o = \exp(-2||o_{target}-o_{object}||)
\end{equation}
$o_{target}$ and $o_{object}$ represent the orientation of the target and the object (in Euler angle, except for the rotation of the bottle around the vertical axis). $R_o$ represents the reward for reorientation.
\begin{equation}
    R_r = \exp(-10||p_{palm}-p_{object}||)
\end{equation}
$p_{palm}$ and $p_{object}$ represent the positions of the palm of the arm model and the position of object. $R_r$ encourages the model to grab the object.

The state space consists of simulation time, joint positions, joint velocities, muscle forces, muscle lengths, muscle velocities and muscle activations. The task-specific observations include the positions of the object and the target, the orientations of the object and the target, the error of position and orientation, the position of the model's palm and the distance between the palm and the object.

\textbf{MyoHand-Reorient100} \quad \textit{MyoHand} model in MyoSuite with 23 DoF and 39 muscles needs to rotate a set of 4 objects, each with 25 different geometries, to a given orientation without dropping them. This set of 100 objects is randomly presented, and initialized onto the hand. The reward function is:
\begin{equation}
    R = -w_pR_p+w_oR_o-w_dR_d-w_aR_a+w_bR_b
\end{equation}
$R_a = ||act||/N_a$, promotes efficient actuator usage. The weights $w_p$, $w_o$, $w_d$, $w_a$ and $w_b$ determine the importance of each reward term. The other terms are as follow:
\begin{equation}
    R_p = ||p_{target}-p_{object}||
\end{equation}
$p_{target}$ and $p_{object}$ represent the positions of the target and the object. $R_p$ keeps the object at its initial position (i.e. onto the palm).
\begin{equation}
    R_o = \cos (o_{target}-o_{object})
\end{equation}
$o_{target}$ and $o_{object}$ represent the orientations of the target and the object (in vector). $R_o$ represents the reward for reorientation.
\begin{equation}
    R_d = (||p_{target}-p_{object}||>0.075)
\end{equation}
$R_d$ represents the penalty for dropping objects.
\begin{align}
    R_b &= (\cos (o_{target}-o_{object}) > 0.9) \times (||p_{target}-p_{object}||<0.075) \notag \\ &+ 5\times(\cos (o_{target}-o_{object}) > 0.95)\times(||p_{target}-p_{object}||<0.075) 
\end{align}
$R_b$ represents the bonus reward for simultaneous rotational and positional alignment above a threshold.

The state space consists of simulation time, joint positions, joint velocities, muscle forces, muscle lengths, muscle velocities and muscle activations. The task-specific observations include the positions of the object and the target, the orientations of the object and the target, the velocities of objects, and the error of position and orientation.

\textbf{Ostrich-Run} \quad An \textit{Ostrich} model \cite{la2021ostrichrl} with 50 joints actuated by 120 muscles is used. It needs to run horizontally as fast as possible, rewarded only by the velocity of the center of mass projected to the x-axis. 
\begin{equation}
    R = w_v v_x^{COM}
\end{equation}

The state space consists of joint positions (except $x$ position for the base segment), joint velocities, muscle forces, muscle lengths, muscle velocities and muscle activations. The task-specific observations include the height of ostrich torso, the height of the feet and the horizontal velocity.

\renewcommand{\thetable}{A.1}
\begin{table}[htbp]
\caption{The action, state dimensions, number of joints and episode length of all the environments.}
\label{env_tab}
\vskip 0.15in
\begin{center}
\begin{tabular}{lcccc}
\toprule
Environment & Action dimensions & State dimensions & Number of joints & Episode length\\
\midrule
FullBody-Gait  & 700 & 2971 & 206 (6 for the base) & 3000 \\
Legs-Walk & 100 & 488 & 36 (6 for the base) & 1000 \\
MyoLegs-Walk & 80 & 403 & 34 (6 for the base) & 1000 \\
Arm-Locate & 81 & 442 & 48 (6 for the object) & 200 \\
MyoHand-Reorient100 & 39 & 200 & 29 (6 for the object) & 50 \\
Ostrich-Run & 120 & 596 & 56 (6 for the base) & 1000 \\
\bottomrule
\end{tabular}
\end{center}
\vskip -0.1in
\end{table}

\section{Implementation Details} 

\subsection{Action normalization}

Our preliminary experiments reveal that in MyoSuite, the action space, originally [0, 1], is normalized to [-1, 1] using Equation \ref{eq:norm}. This normalization method enhances training effectiveness. Consequently, we apply this normalization approach in all our environments and algorithm comparison experiments.

\begin{equation}
\label{eq:norm}
    a = \frac{1}{1+e^{-5(a-0.5)}}
\end{equation}

\subsection{Dynamical synergistic representation generation} \label{group_num}

\begin{figure}[htbp]
\centering
\subfigure[] {
\label{fig:kmedoids_analysis}
\includegraphics[width=0.5\columnwidth]{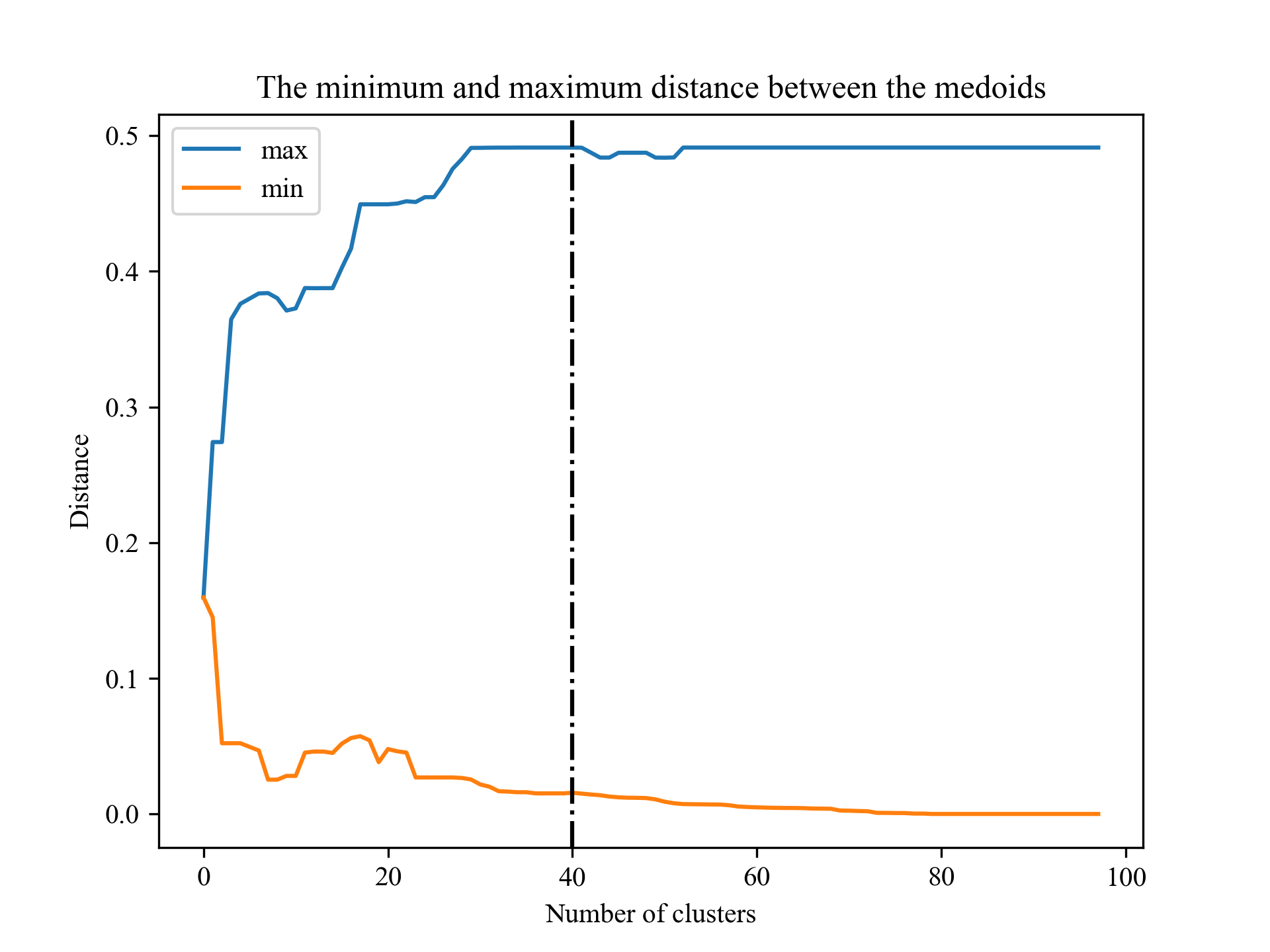}  
} 
\subfigure[] { 
\label{fig:grouping_b}     
\includegraphics[width=0.46\columnwidth]{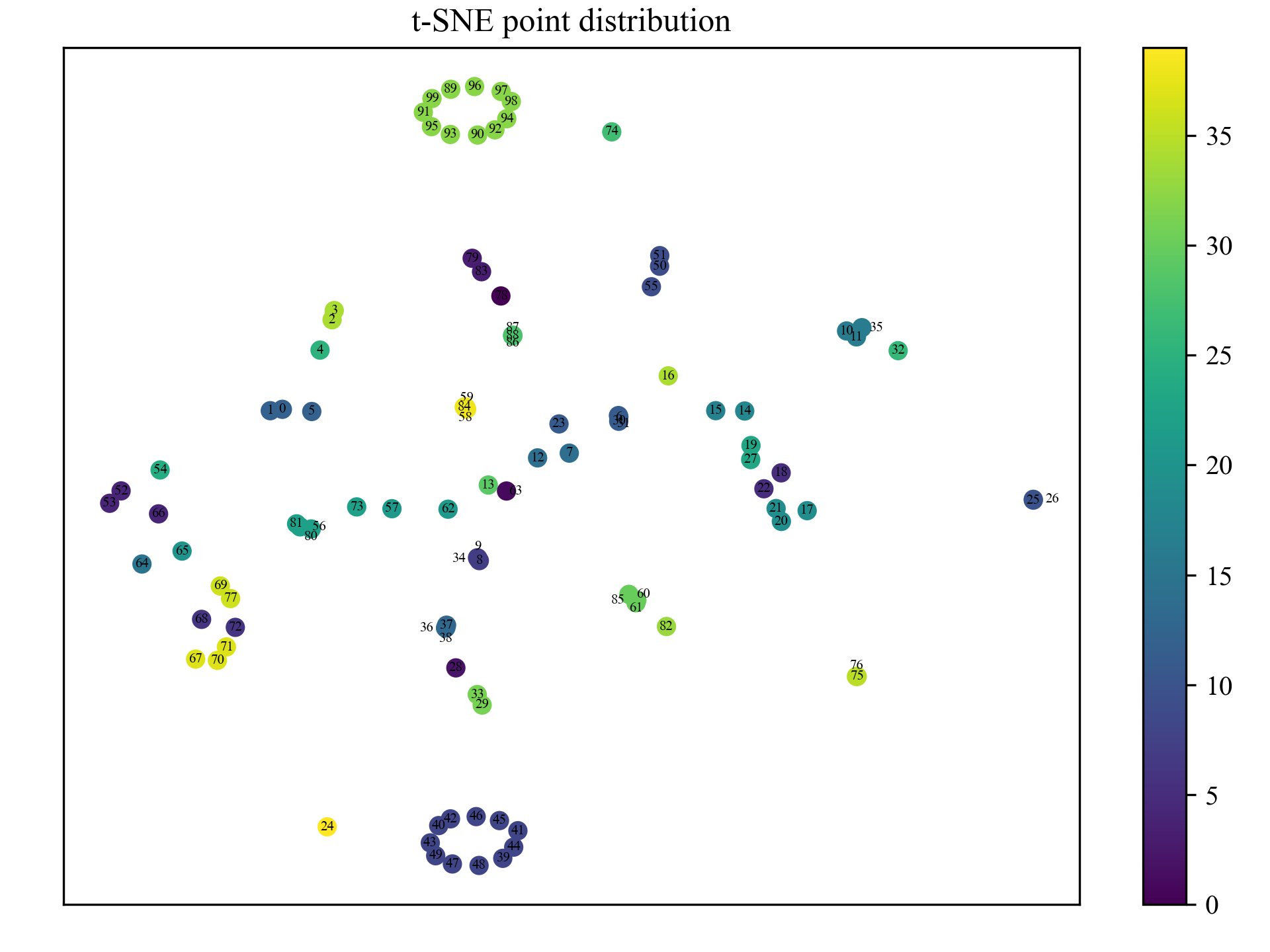}     
}    
\caption{\textbf{Generation criteria and visualization.} (a) In \textit{Legs} model, the K-medoids algorithm is employed for clustering with varying cluster numbers. The curve depicting the maximum and minimum distance between cluster centers changes with the number of clusters. (b) t-SNE algorithm is used to reduce the cosine similarity distance to two dimensions for visualization.}     
\label{fig:grouping}     
\end{figure}

In the process of representation generation, determining the number of groups is crucial. In Figure \ref{fig:kmedoids_analysis}, we illustrate the maximum and minimum values of the distance among cluster centers for different group configurations. The algorithm exhibits robust and explainable performance when we choose an appropriate number of clusters where the difference between maximum and minimum distances are large enough. In Figure \ref{fig:grouping_b}, it is evident that when the selected number of groups is 40, the t-SNE visualization maintains symmetry and interpretability.

\begin{figure}[htbp]
    \centering
    \includegraphics[width=1\linewidth]{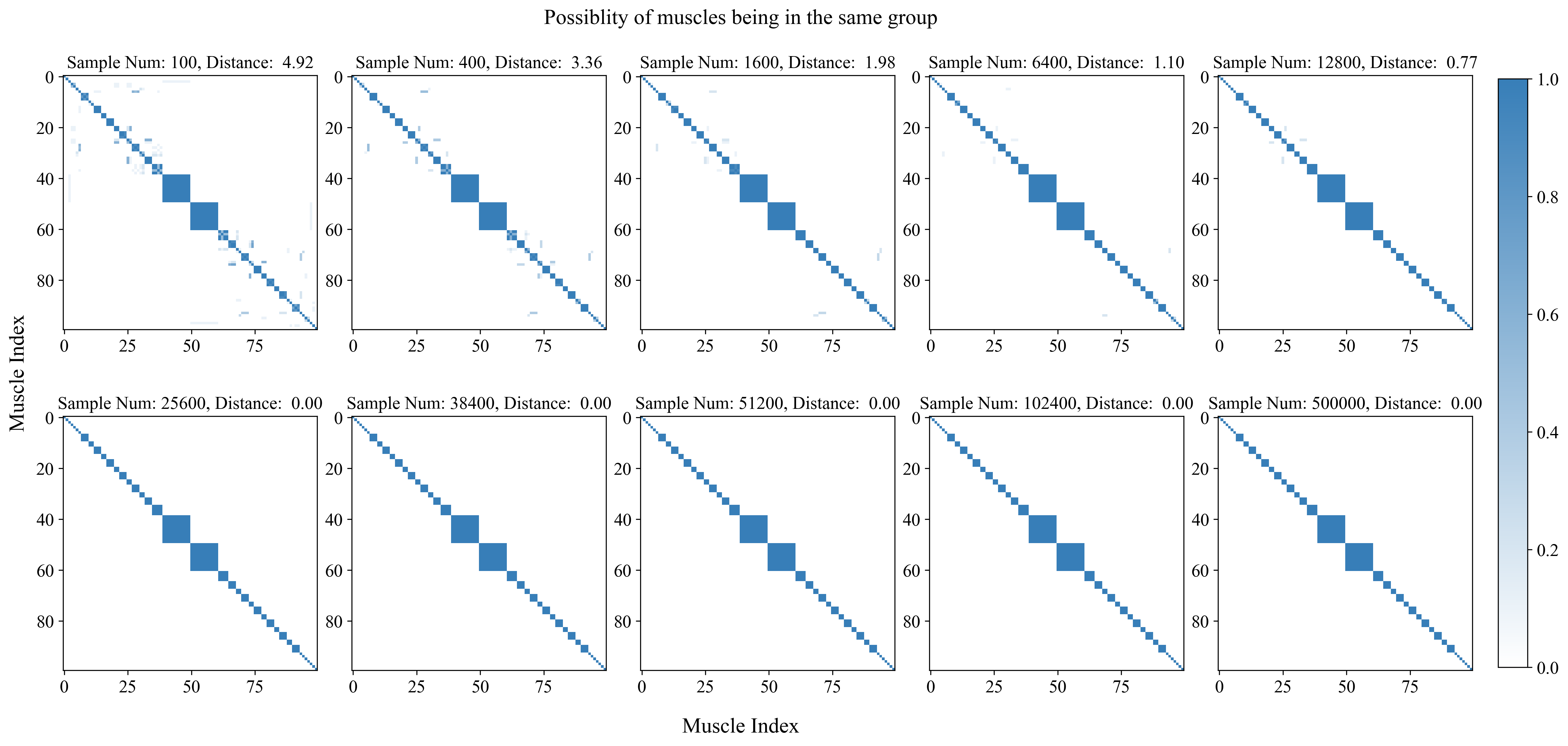}
    \vskip -0.1in
    \caption{Grouping matrix $P_{n}$ derived from varying sample sizes, and the distance (Frobenius norm) between them and a 500K-sample grouping matrix. $p_{ij}$ is the probability that muscle $i$ and muscle $j$ are in the same group (averaged over 10 seeds). The grouping results are shown to converge to their final grouping with a data point quantity as low as 25,600. Even if we have only 100 data points, the grouping result is similar to the final result.}
    \label{fig:compare}
\end{figure}

\begin{figure}[htbp]
    \centering
    \includegraphics[width=0.5\linewidth]{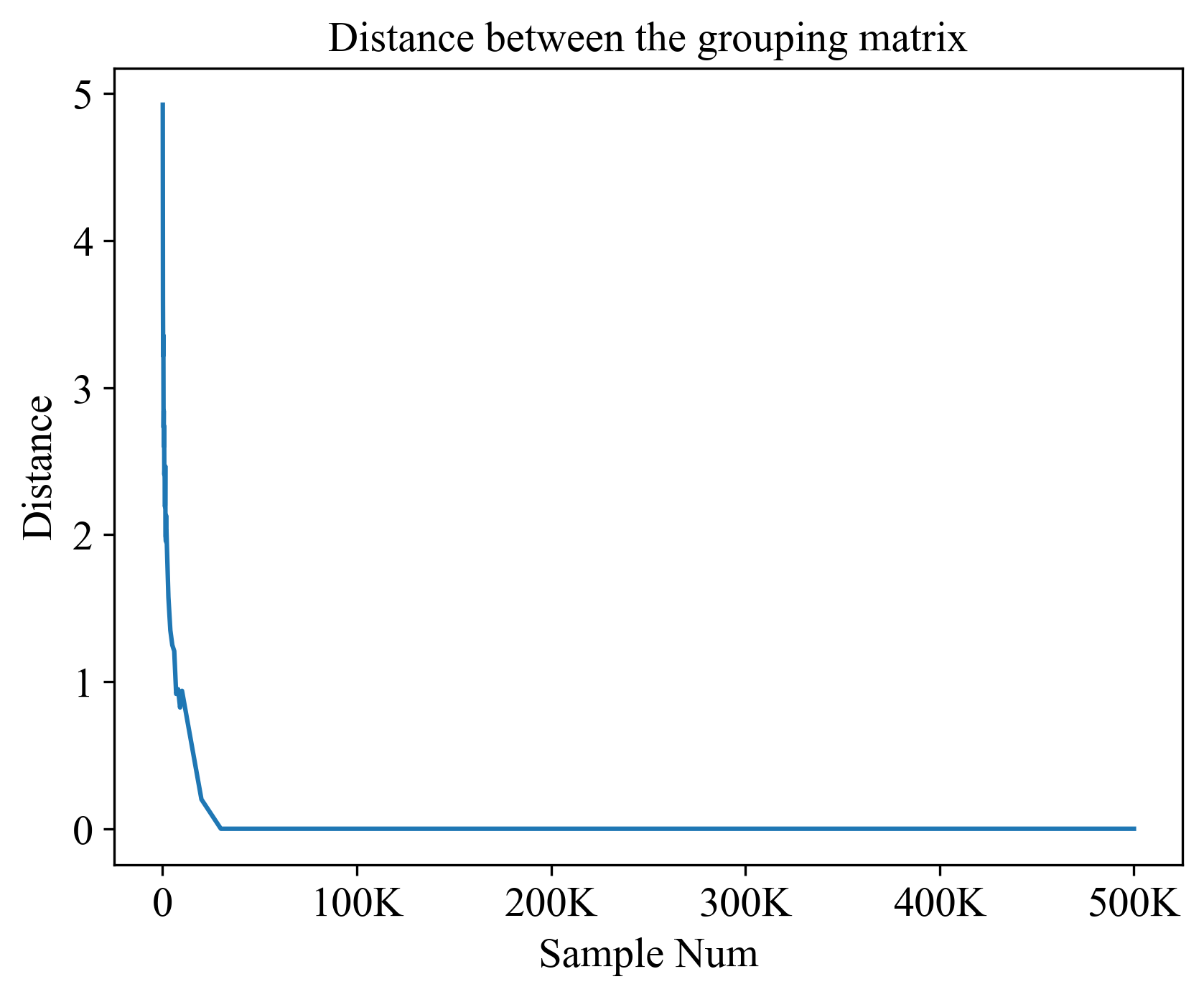}
    \vskip -0.1in
    \caption{The distance (Frobenius norm) between grouping matrices derived from varying sample sizes and a 500K-sample grouping matrix.}
    \label{fig:distance}
\end{figure}

As illustrated in Figure \ref{fig:compare} and Figure \ref{fig:distance}, the grouping results are shown to converge to their final grouping with a data point quantity as low as 25,600. It's also displayed that even if we have only 100 data points, the grouping result is similar to the final result.

\subsection{Baselines}

We compare our algorithm with the best performing baselines, SAC, SAR and DEP-RL. SAR and DEP-RL algorithms are implemented using the official released code. SAC, SAR and DynSyn all adopt the DRL framework Stable baselines3, and DEP-RL adopts the Tonic framework. All algorithms adopt SAC as the basic algorithm. The specific algorithm parameters of SAR and DEP-RL are those reported in the original papers, and for models with similar complexity we use the same parameters. See Table \ref{tab:algo_para_1} and Table \ref{tab:algo_para_2} for details.

\subsection{Hyperparameters of \algo\ and baselines} \label{algo_detail}

Algorithm hyperparameters are summarized in Table \ref{tab:algo_para_1} and \ref{tab:algo_para_2}.

\renewcommand{\thetable}{A.2}
\begin{table}[h]
\caption{The task and reward parameters of all the environments.}
\label{env_para}
\begin{center}
\begin{tabular}{ccccc}
\toprule
Task & FullBody-Gait &  & Ostrich-Run &  \\ \hline
 & Parameter & Value & Parameter & Value \\ \hline
 & Min pelvis height & 0.6 & Min head height & 0.9 \\
 & Min sternum height & 1 & Min pelvis height & 0.6 \\
 & Max rot. & 0.8 & Min torso angle & -0.8 \\
 &  &  & Max torso angle & 0.8 \\ \hline
 Reward & Pos. & 1 & Vel. & 1 \\
 & Vel. & 0.005 &  &  \\
 & Concerned & 1 &  &  \\
 & Done & 10 &  &  \\ \hline \hline
Task & Legs-Walk &  & Arm-Locate &  \\ \hline
 & Parameter & Value & Parameter & Value \\  \hline
 & Min height & 0.8 & Pos. x target bound & (-0.1, 0.1) \\
 & Max rot. & 0.8 & Pos. y target bound & (-0.2, 0.2) \\
 & Hip period & 100 & Pos. z target bound & (0.1, 0.3) \\ \hline
 Reward & Target forward vel. & 1.2 (3 in Fast env) & Ori. x target bound & (-0.4, 0.4) \\
 & Target lateral vel. & 0 (1.2 in Diagnal env) & Ori. y target bound & (-0.4, 0.4) \\
 & Vel. & 5 & Pos. & 50 \\
 & Cyclic hip & 10 & Ori. & 5 \\
 & Ref rot. & 10 & Reach & 10 \\
 & Joint angle & 5 & Action reg. & 1 \\
 & Done & 100 &  &  \\ \hline \hline
Task & MyoLegs-Walk &  & MyoHand-Reorient100 &  \\ \hline
 & Parameter & Value & Parameter & Value \\ \hline
 & Min height & 0.8 & - & - \\
 & Max rot. & 0.8 &  &  \\
 & Hip period & 100 &  &  \\
 & Target forward vel & 1.2 &  &  \\
 & Target lateral vel & 0 &  &  \\
 & Terrain & \makecell[c]{Flat (Rough, Hilly, \\in separate envs)} &  &  \\ \hline
 Reward & Vel. & 5 & Pos. & 1 \\
 & Cyclic hip & 10 & Ori. & 1 \\
 & Ref rot. & 10 & Drop & 5 \\
 & Joint angle & 5 & Action reg. & 5 \\
 & Action reg. & 1 & Bonus & 10 \\
 & Done & 100 &  & \\
\bottomrule
\end{tabular}
\end{center}
\end{table}

\renewcommand{\thetable}{A.3}
\begin{table}[h]
\centering
\caption{Parameters of SAC, \algo, SAR and DEP-RL in the standard tasks}
\label{tab:algo_para_1}
\begin{tabular}{|c|c|cccccc|}
\hline
\multirow{2}{*}{Algorithm} & \multirow{2}{*}{Parameter} & \multicolumn{6}{c|}{Task} \\ \cline{3-8} 
 &  & \multicolumn{1}{c|}{\begin{tabular}[c]{@{}c@{}}FullBody-\\ Gait\end{tabular}} & \multicolumn{1}{c|}{\begin{tabular}[c]{@{}c@{}}Legs-\\ Walk\end{tabular}} & \multicolumn{1}{c|}{\begin{tabular}[c]{@{}c@{}}MyoLegs-\\ Walk\end{tabular}} & \multicolumn{1}{c|}{\begin{tabular}[c]{@{}c@{}}Arm-\\ Locate\end{tabular}} & \multicolumn{1}{c|}{\begin{tabular}[c]{@{}c@{}}MyoHand-\\ Reorient100\end{tabular}} & \begin{tabular}[c]{@{}c@{}}Ostrich-\\ Run\end{tabular} \\ \hline
\multirow{16}{*}{SAC} & Learning rate & \multicolumn{6}{c|}{linear\_schedule(0.001)} \\ \cline{2-8} 
 & Batch size & \multicolumn{6}{c|}{256} \\ \cline{2-8} 
 & Buffer size & \multicolumn{6}{c|}{3e6} \\ \cline{2-8} 
 & Warmup steps & \multicolumn{6}{c|}{100} \\ \cline{2-8} 
 & Discount factor & \multicolumn{6}{c|}{0.98} \\ \cline{2-8} 
 & Soft update coeff. & \multicolumn{6}{c|}{2} \\ \cline{2-8} 
 & Train frequency (steps) & \multicolumn{6}{c|}{1} \\ \cline{2-8} 
 & Gradient steps & \multicolumn{6}{c|}{4} \\ \cline{2-8} 
 & Traget update interval & \multicolumn{6}{c|}{1} \\ \cline{2-8} 
 & Environment number & \multicolumn{6}{c|}{80} \\ \cline{2-8} 
 & Entropy coeff. & \multicolumn{6}{c|}{auto} \\ \cline{2-8} 
 & Target entropy & \multicolumn{6}{c|}{auto} \\ \cline{2-8} 
 & Policy hiddens & \multicolumn{1}{c|}{{[}512, 300{]}} & \multicolumn{5}{c|}{{[}256, 256{]}} \\ \cline{2-8} 
 & Q hiddens & \multicolumn{1}{c|}{{[}512, 300{]}} & \multicolumn{5}{c|}{{[}256, 256{]}} \\ \cline{2-8} 
 & Activation & \multicolumn{6}{c|}{ReLU} \\ \cline{2-8} 
 & Training steps & \multicolumn{1}{c|}{5e7} & \multicolumn{4}{c|}{3e6} & \multicolumn{1}{c|}{5e6} \\ \hline
\multirow{6}{*}{\algo} & Control Amplitude & \multicolumn{1}{c|}{5} & \multicolumn{1}{c|}{10} & \multicolumn{1}{c|}{10} & \multicolumn{1}{c|}{5} & \multicolumn{1}{c|}{100} & 10 \\ \cline{2-8} 
 & Trajectory steps & \multicolumn{6}{c|}{5e5} \\ \cline{2-8} 
 & Control frequency & \multicolumn{6}{c|}{10} \\ \cline{2-8} 
 & Number of groups & \multicolumn{1}{c|}{100} & \multicolumn{2}{c|}{40} & \multicolumn{2}{c|}{25} & 40 \\ \cline{2-8} 
 & $a_{D}$ & \multicolumn{1}{c|}{3e7} & \multicolumn{5}{c|}{1e6} \\ \cline{2-8} 
 & $k_{D}$ & \multicolumn{1}{c|}{5e-9} & \multicolumn{5}{c|}{5e-8} \\ \hline
\multirow{3}{*}{SAR} & Dimensionality & \multicolumn{1}{c|}{200} & \multicolumn{5}{c|}{20} \\ \cline{2-8} 
 & Blend weight & \multicolumn{6}{c|}{0.66} \\ \cline{2-8} 
 & \begin{tabular}[c]{@{}c@{}}Training steps\\ (play phase + training phase)\end{tabular} & \multicolumn{1}{c|}{2e7+3e7} & \multicolumn{4}{c|}{1.5e6+1.5e6}  & \multicolumn{1}{c|}{2e6+3e6}  \\ \hline
\multirow{18}{*}{DEP-RL} & Bias rate & \multicolumn{5}{c|}{0.002} & 0.03 \\ \cline{2-8} 
 & Buffer size of DEP & \multicolumn{5}{c|}{200} & 90 \\ \cline{2-8} 
 & Intervention length & \multicolumn{5}{c|}{5} & 4 \\ \cline{2-8} 
 & Intervention proba & \multicolumn{6}{c|}{0.0004} \\ \cline{2-8} 
 & Kappa & \multicolumn{5}{c|}{1169.7} & 20 \\ \cline{2-8} 
 & Normalization & \multicolumn{6}{c|}{Independent} \\ \cline{2-8} 
 & Q norm selector & \multicolumn{6}{c|}{l2} \\ \cline{2-8} 
 & regularization & \multicolumn{6}{c|}{32} \\ \cline{2-8} 
 & s4avg & \multicolumn{5}{c|}{2} & 1 \\ \cline{2-8} 
 & Sensor delay & \multicolumn{6}{c|}{1} \\ \cline{2-8} 
 & tau & \multicolumn{5}{c|}{40} & 8 \\ \cline{2-8} 
 & Test episode every & \multicolumn{6}{c|}{3} \\ \cline{2-8} 
 & Time dist & \multicolumn{6}{c|}{5} \\ \cline{2-8} 
 & With learning & \multicolumn{6}{c|}{True} \\ \cline{2-8} 
 & Return steps & \multicolumn{6}{c|}{2} \\ \cline{2-8} 
 & Entropy coeff. & \multicolumn{6}{c|}{0.2} \\ \cline{2-8} 
 & Learning rate & \multicolumn{6}{c|}{3e-4} \\ \cline{2-8} 
 & \begin{tabular}[c]{@{}c@{}}Environment number\\ (parallel * sequential)\end{tabular} & \multicolumn{6}{c|}{80*1} \\ \hline
\end{tabular}
\end{table}

\renewcommand{\thetable}{A.4}
\begin{table}[h]
\centering
\caption{Parameters of SAC, \algo, SAR and DEP-RL in the generalization tasks}
\label{tab:algo_para_2}
\begin{tabular}{|c|c|cccc|}
\hline
\multirow{2}{*}{Algorithm} & \multirow{2}{*}{Parameter} & \multicolumn{4}{c|}{Task} \\ \cline{3-6} 
 &  & \multicolumn{1}{c|}{\begin{tabular}[c]{@{}c@{}}Legs-\\ Walk-\\ Diagnal\end{tabular}} & \multicolumn{1}{c|}{\begin{tabular}[c]{@{}c@{}}Legs-\\ Walk-\\ Fast\end{tabular}} & \multicolumn{1}{c|}{\begin{tabular}[c]{@{}c@{}}MyoLegs-\\ Walk-\\ Hilly\end{tabular}} & \begin{tabular}[c]{@{}c@{}}MyoLegs-\\ Walk-\\ Rough\end{tabular} \\ \hline
\multirow{16}{*}{SAC} & Learning rate & \multicolumn{4}{c|}{linear\_schedule(0.001)} \\ \cline{2-6} 
 & Batch size & \multicolumn{4}{c|}{256} \\ \cline{2-6} 
 & Buffer size & \multicolumn{4}{c|}{3e6} \\ \cline{2-6} 
 & Warmup steps & \multicolumn{4}{c|}{100} \\ \cline{2-6} 
 & Discount factor & \multicolumn{4}{c|}{0.98} \\ \cline{2-6} 
 & Soft update coeff. & \multicolumn{4}{c|}{2} \\ \cline{2-6} 
 & Train frequency (steps) & \multicolumn{4}{c|}{1} \\ \cline{2-6} 
 & Gradient steps & \multicolumn{4}{c|}{4} \\ \cline{2-6} 
 & Traget update interval & \multicolumn{4}{c|}{1} \\ \cline{2-6} 
 & Environment number & \multicolumn{4}{c|}{80} \\ \cline{2-6} 
 & Entropy coeff. & \multicolumn{4}{c|}{auto} \\ \cline{2-6} 
 & Target entropy & \multicolumn{4}{c|}{auto} \\ \cline{2-6} 
 & Policy hiddens & \multicolumn{4}{c|}{{[}256, 256{]}} \\ \cline{2-6} 
 & Q hiddens & \multicolumn{4}{c|}{{[}256, 256{]}} \\ \cline{2-6} 
 & Activation & \multicolumn{4}{c|}{ReLU} \\ \cline{2-6} 
 & Training steps & \multicolumn{4}{c|}{3e6} \\ \hline
\multirow{2}{*}{\algo} & $a_{D}$ & \multicolumn{4}{c|}{1e6} \\ \cline{2-6} 
 & $k_{D}$ & \multicolumn{4}{c|}{5e-6} \\ \hline
\multirow{3}{*}{SAR} & Dimensionality & \multicolumn{4}{c|}{20} \\ \cline{2-6} 
 & Blend weight & \multicolumn{4}{c|}{0.66} \\ \cline{2-6} 
 & \begin{tabular}[c]{@{}c@{}}Training steps\\ (play phase + training phase)\end{tabular} & \multicolumn{4}{c|}{0+3e6} \\ \hline
\multirow{18}{*}{DEP-RL} & Bias rate & \multicolumn{4}{c|}{0.002} \\ \cline{2-6} 
 & Buffer size of DEP & \multicolumn{4}{c|}{200} \\ \cline{2-6} 
 & Intervention length & \multicolumn{4}{c|}{5} \\ \cline{2-6} 
 & Intervention proba & \multicolumn{4}{c|}{0.0004} \\ \cline{2-6} 
 & Kappa & \multicolumn{4}{c|}{1169.7} \\ \cline{2-6} 
 & Normalization & \multicolumn{4}{c|}{Independent} \\ \cline{2-6} 
 & Q norm selector & \multicolumn{4}{c|}{l2} \\ \cline{2-6} 
 & regularization & \multicolumn{4}{c|}{32} \\ \cline{2-6} 
 & s4avg & \multicolumn{4}{c|}{2} \\ \cline{2-6} 
 & Sensor delay & \multicolumn{4}{c|}{1} \\ \cline{2-6} 
 & tau & \multicolumn{4}{c|}{40} \\ \cline{2-6} 
 & Test episode every & \multicolumn{4}{c|}{3} \\ \cline{2-6} 
 & Time dist & \multicolumn{4}{c|}{5} \\ \cline{2-6} 
 & With learning & \multicolumn{4}{c|}{True} \\ \cline{2-6} 
 & Return steps & \multicolumn{4}{c|}{2} \\ \cline{2-6} 
 & Entropy coeff. & \multicolumn{4}{c|}{0.2} \\ \cline{2-6} 
 & Learning rate & \multicolumn{4}{c|}{3e-4} \\ \cline{2-6} 
 & \begin{tabular}[c]{@{}c@{}}Environment number\\ (parallel $\times$ sequential)\end{tabular} & \multicolumn{4}{c|}{80$\times$1} \\ \hline
\end{tabular}
\end{table}